\documentclass[twoside,11pt]{article}
\usepackage{jair, rawfonts}
\PassOptionsToPackage{table}{xcolor}
\usepackage[normalem]{ulem}
\usepackage{amsfonts}
\usepackage{url,lineno,microtype,subcaption}
\usepackage{caption}
\usepackage[export]{adjustbox}
\usepackage{makecell}
\usepackage{floatrow}
\usepackage{graphicx}
\usepackage{xparse}
\usepackage[linguistics, edges]{forest}
\usepackage{tikz}
\usetikzlibrary{calc,positioning}
\usetikzlibrary{decorations, decorations.text}
\usepackage{array}
\usepackage{tabularx}
\usepackage[ruled,vlined]{algorithm2e}
\SetKwInput{KwInput}{Input}
\SetKwInput{KwOutput}{Output}
\usepackage{hyperref}
\usepackage[round]{natbib}

\DeclareGraphicsExtensions{.pdf,.jpeg,.png,.jpg,.eps}

\usepackage{xspace}
\makeatletter

\DeclareRobustCommand\onedot{\futurelet\@let@token\@onedot}
\def\@onedot{\ifx\@let@token.\else.\null\fi\xspace}

\def\eg{\emph{e.g}\onedot}

\def\etc{\emph{etc}\onedot} 

\def\wrt{w.r.t\onedot}

\makeatother

\usepackage{amsmath, amssymb}
\usepackage{cleveref}
\usepackage{multirow}

\newcommand{\real}{\mathbb{R}}
\newcommand{\mdp}{\mathcal{M}}
\newcommand{\states}{\mathcal{S}}
\newcommand{\actions}{\mathcal{A}}
\newcommand{\transitions}{T}
\newcommand{\rewards}{\mathcal{R}}
\newcommand{\initstate}{d_0}
\newcommand{\discount}{\gamma}

\newcommand{\observations}{\mathcal{Z}}
\newcommand{\belief}{\mathcal{B}}
\newcommand{\numtasks}{\mathcal{N}}
\newcommand{\observationFunction}{\mathcal{O}}
\newcommand{\bs}{\mathbf{s}}
\newcommand{\ba}{\mathbf{a}}
\newcommand{\bz}{\mathbf{z}}
\newcommand{\task}{\texttt{T}}
\newcommand{\taskmath}{\mathcal{T}}
\newcommand{\policy}{\pi}
\newcommand{\return}{\mathcal{G}}
\newcommand{\data}{\mathcal{X}}
\newcommand{\labels}{\mathcal{Y}}

\definecolor{LightCyan}{rgb}{0.88,1,1}
\usepackage{tabulary,booktabs}
\usepackage{ragged2e}
\newcolumntype{L}[1]{>{\hsize=#1\hsize\RaggedRight} X}

\newtheorem{remark}{Remark}

\newlength{\layerwd}
\newcounter{outermost}

\NewDocumentEnvironment{onion}{sm}{%*= draw axes; #1: thickness of each annulus
    \begin{tikzpicture}
        \setlength{\layerwd}{#2}%
        \setcounter{outermost}{1}
        \IfBooleanT{#1}{%
            \draw[<->] (-4,0) -- (4,0);
            \draw[<->] (0,4) -- (0,-4);
        }%
}{%
    \foreach \A in {1,...,\theoutermost}{\draw[thick] (0,0) circle (\A*\layerwd+\layerwd);}
    \end{tikzpicture}
}

\NewDocumentCommand{\annulus}{sO{white}mmmo}{%
    \filldraw[thick,fill=#2] (#4:#3*\layerwd) %% start here
        arc [radius=#3*\layerwd, start angle=#4, delta angle=#5-#4] %% inner arc
        -- (#5:#3*\layerwd+\layerwd) %% move out
        arc [radius=#3*\layerwd+\layerwd, start angle=#5, delta angle=#4-#5] %% outer arc
        -- cycle; %% Back to the beginnning
    \pgfmathsetmacro{\tmp}{(#5-#4)/2 +#4} %% Locate the middle of the arc
    \IfNoValueF{#6}{%
        \IfBooleanTF{#1}
        {%
            \begingroup
                \pgfmathsetmacro{\rpTF}{ifthenelse(\tmp>180,"false","true")}
                \def\\{\space} %% A safety precaution, \\ = space on decorated text
                \path[rotate=\tmp-180,postaction={
                    decorate,
                    decoration={
                        text along path,
                        raise=-3pt,
                        text align={align=center},
                        reverse path=\rpTF,
                        text=#6
                    }
                }] (0,0) circle (#3*\layerwd+0.5*\layerwd);
            \endgroup
        }%% 
        {%
            \pgfmathsetmacro{\rpTF}{ifthenelse(\tmp>180,\tmp+90,\tmp-90)}
            \node[inner sep=0pt, %%% If there is text, print it
            text width=#3*\layerwd*3+\layerwd,
            align=center,
            rotate=\rpTF,
            font=\footnotesize] at (\tmp:#3*\layerwd+0.5*\layerwd)
            {#6};
        }%
    }%
    \ifnum\theoutermost<#3\setcounter{outermost}{#3}\fi
}

\usepackage{bm}
\usepackage[mathscr]{euscript}

\def\1{\bm{1}}

\def\vtheta{{\bm{\theta}}}

\def\vx{{\bm{x}}}
\def\vy{{\bm{y}}}

\DeclareMathAlphabet{\mathsfit}{\encodingdefault}{\sfdefault}{m}{sl}
\SetMathAlphabet{\mathsfit}{bold}{\encodingdefault}{\sfdefault}{bx}{n}

\newcommand{\E}{\mathop{\mathbb{E}}\nolimits}

\newcommand{\R}{\mathbb{R}}

\DeclareMathOperator*{\argmin}{arg\,min}

\begin{document}

\title{Advancements and Challenges in Continual Reinforcement Learning: A Comprehensive Review}

\author{\name Amara Zuffer \email fathima.mohamedzuffer@monash.edu \\
        \addr Electrical and Computer Science Engineering Department, \
       Faculty of Engineering, Monash University, Clayton, Victoria, Australia
        \AND
        \name Michael Burke \email michael.g.burke@monash.edu \\
       \addr Electrical and Computer Science Engineering Department, \
       Faculty of Engineering, Monash University, Clayton, Victoria, Australia
       \AND
       \name Mehrtash Harandi \email mehrtash.harandi@monash.edu \\
       \addr Electrical and Computer Science Engineering Department, \
       Faculty of Engineering, Monash University, Clayton, Victoria, Australia}

% For research notes, remove the comment character in the line below.
% \researchnote

\maketitle

\begin{abstract}
The diversity of tasks and dynamic nature of reinforcement learning (RL) require RL agents to be able to learn sequentially and continuously, a learning paradigm known as continuous reinforcement learning.  
% The dynamic landscape of Reinforcement Learning has lead the emergence of Continual Reinforcement Learning, a pivotal approach addressing challenges in the sequential learning of evolving environments and diverse task distributions. 
This survey reviews how continual learning transforms RL agents into dynamic continual learners. This enables RL agents to acquire and retain useful and reusable knowledge seamlessly. The paper delves into fundamental aspects of continual reinforcement learning, exploring key concepts, significant challenges, and novel methodologies. Special emphasis is placed on recent advancements in continual reinforcement learning within robotics, along with a succinct overview of evaluation environments utilized in prominent research, facilitating accessibility for newcomers to the field. The review concludes with a discussion on limitations and promising future directions, providing valuable insights for researchers and practitioners alike.
\end{abstract}

\section{Introduction}

% The paper focuses on discussing:
% \begin{enumerate}
%     \item Introduction into CL, RL, CRL, lifelong CL - mathematical pending
%     \item Problem definition -done
%     \item Assumptions
%     \item A breakdown of current methods belonging to each category of CL -done
%     \item Model-based vs model-free learning \AZ{DONE}
%     \item Transformer based approaches including prompts
%     \item distinguish CL and lifelong be
% tween other forms of learning
%     \item Task-aware, task-agnostic (train and eval) -done
%     \item How each RL algorithm is improved/ how they contribute for CL. Any algorithmic enhancements introduced.
%     \item Why haven't some on-policy and off-policy RL algorithms not been tested for CL?
%     \item Would the change in task order break the fine tuned result?
%     \item Benchmarks: The RL environments used for these studies (MDP, POMDP) -done
%     \item CNN based vs non CNN based input processing
%     \item Metrics
%     \item Comparisons
%     \item Discussion, Future directions, issues - done
%     \item overview of algorithm for regular, param isola etc.
%     \item in 2.1 explain the connection to the brain and then come to CF. 2.2 explain how to overcome CF using algorithms
% \end{enumerate}
% \textbf{Notes}
% \cite{xie2020deep} has been tested on Mujoco via introducing non-stationarity by changing target velocities and wind forces.
% \cite{sutton2018reinforcement}

Learning and adaptability observed in the human mind/brain~\citep{bassett2011dynamic, hassabis2017neuroscience, soltani2019adaptive} have inspired numerous advancements in the field of machine learning and AI~\citep{krizhevsky2012imagenet, lecun2015deep, mnih2015human,  silver2017mastering}. 
% These models strive to replicate the learning behavior of humans when needed. 
% Hence, their success is often compared to the human intelligence baselines. 
One of the most vital capabilities of us as human beings is the ability to learn sequentially and in a continual manner throughout our lifetime. This capability enables us to adapt amidst changes in our surroundings and experiences. Continual Learning (CL) is the study of the ongoing acquisition of knowledge, contributing to the gradual development of more complex behaviors while retaining previously learned experiences. 
% Past experiences are leveraged to improve and expedite the knowledge acquisition of future tasks. 
Traditionally,  machine learning models are trained to be specialized in solving only one specific problem (\eg, classifying felids) and require training from scratch in the event of the need to learn new tasks (\eg, learning to classify canids in addition to felids).
% Lifelong Learning is another related form of learning that assumes the possibility of acquiring knowledge, skills and experiences throughout the entire lifelong span of individuals or systems not restricted by the concept of tasks. In contrast to this in CL a task based learning approach is assumed 
CL is instrumental in enabling a single model to handle sequential learning problems of multiple tasks and has been explored under supervised~\citep{zenke2017continual, parisi2019continual,  hadsell2020embracing, de2021continual, mcdonnell2024ranpac} and unsupervised~\citep{rao2019continual, taufique2022unsupervised, davari2022probing, chawla2024continual, gomez2024plasticity} settings extensively in the recent years. A major difficulty in learning continually in AI is to address Catastrophic Forgetting (CF) caused by adapting the model (\eg, a Neural Network (NN)) to excel at the latest tasks, leading the model to forget previously learned tasks.
% \MH{first define CF here. Also, we need to emphasize that we are interested in learning with NNs, so below, when you say RL, CF does not apply unless we consider Deep RL}

In this survey, we are chiefly interested in CL methods for Deep Reinforcement Learning (DRL). {RL involves learning to navigate dynamic or non-stationary environments in order to maximise the expected future reward. Both these environment types experience changes over time, where changes are more predictable in dynamic environments (as they take place within the same environment), whereas changes are unpredictable or irregular in non-stationary environments (as they take place between environments).} The standard RL problem formulation inherently involves continual learning, as the reward can be non-stationary and stochastic, as the state and policy could be. However, current DRL methods are not well-suited for this setting, as deep learning models are primarily designed to handle stationary data distributions. As a result, existing DRL approaches work best in fixed environments that can be exhaustively searched by trial and error like methods, effectively turning learning in these dynamic environments into a stationary problem. This limitation has driven the focus towards continual DRL, which directly addresses these challenges in settings where the assumption of stationarity does not hold or where dynamic environments cannot be simplified.

DRL algorithms, which combine RL with deep neural networks, learn to navigate dynamic environments based on the reward received from the environment, making them particularly well-suited to learn varying experiences continuously. Continual Reinforcement Learning (CRL) extends this framework by developing algorithms capable of learning from non-stationary data distributions while overcoming CF of past tasks~\citep{lesort2020continual}.

Apart from the instabilities DRL algorithms inherently struggle with~\citep{Ding2020ChallengesOR, laskin2020curl}, CRL agents need to,
\begin{enumerate}
    \item Overcome catastrophic forgetting.
    \item Consider the impact of task interference. Learning a new task may disrupt the learning of past or future tasks.
    \item Exhibit knowledge transferability. The agent should be able to reuse its past knowledge to acquire new knowledge efficiently.
    \item Enhance learning efficiency (\eg by employing sample-efficient exploration methods) {to obtain optimal performance in RL tasks while minimizing expensive computation.} 
\end{enumerate}

Recommendation systems~\citep{mi2020ader, zhang2023survey}, language~\citep{biesialska2020continual, winata2023overcoming, zhang2024influential, gamel2024sleepsmart} and robotics~\citep{lesort2020continual, liu2023continual, huang2023improved, gai2024continual} are practical applications of CRL scenarios that demonstrate the need to adapt to dynamic environmental conditions. In recommendation systems, CRL enables the model to continuously learn and update data points such as user preferences and contextual information. These crucial data points isolate each consumer and hence allow the model to make decisions on personalized recommendations. This continuous feedback eventually guides the CL agent to make accurate and timely decisions.
Tasks that involve language, such as dialogue or virtually automated chat systems, function continuously with the involvement of new words, phrases, and expressions that are susceptible to change due to the age group, gender, and context of the user. CRL can assist such AI agents in updating their language understanding and maintaining knowledge from past time periods. Finally, in robotics, CRL empowers robots to adapt to changing environmental conditions, robot dynamics, or even user preferences, by learning from experiences and responses. For example, CRL could also assist to adapt robotic systems to changes in mass or dynamic properties due to wear and tear when completing different tasks~\citep{davchev2022residual}.  Robots that navigate real-world setups should continually update their perception of the environment on obstacles and safety measures\citep{rana2023residual, liu2023ai, wang2023incorporating}.

This survey is intended to act as an entry point to understand RL concepts and extensions to a CRL setting, outlining key concepts, the connection between human cognition and CL, how RL aligns with CL strategies, before providing a breakdown of broad topics and a deep dive into CL and CRL algorithms emphasizing the pros and cons. Finally, we touch upon RL environments used in algorithm performance evaluations, CRL in robotics, evaluation metrics, open challenges, and future directions. Our core goal is to outline why CRL should be focused on as a learning method and what crucial challenges are to be solved in order to improve the performance of existing algorithms, thereby bridging the gap between current CL/RL systems and CRL.

% Paper roadmap:

\paragraph{Scope.}
% \newline
Our aim is to explore the latest advancements in CRL and provide a platform for emerging researchers in the field. While we do not intend to present a taxonomy for this research field, our goal is to provide an easily accessible resource for new researchers. 
% To achieve this, we begin with a brief overview of how CL draws inspiration from the learning procedure of the mammalian brain (\cref{sec:brain_cl}). Subsequently, we offer a comprehensive discussion on CL in~\cref{sec:CL}, highlighting its distinctions from other learning paradigms, addressing challenges, and reviewing the most recent and notable work in the sector. In~\cref{sec:RL}, we delve into RL by examining breakdowns and sub-segments, particularly focusing on learning environments and algorithms.
To achieve this, we begin with a comprehensive overview of mammalian learning procedures followed by CL and RL in \cref{sec:background}. We provide a detailed exploration of CRL in~\cref{sec:CRL}, including distinct categorizations, and further discuss CRL approaches in robotics in \cref{sec:CRL_robo}. Then we touch upon the latest metrics used to evaluate these CRL algorithms in both non-robotic and robotic CL scenarios in~\cref{sec:metrics}. We wrap up by discussing contemporary challenges and open problems in the field in~\cref{sec:future}.

\section{Background}
\label{sec:background}

In this section, we begin with a brief overview of how CL draws inspiration from the learning procedure of the mammalian brain (\cref{sec:brain_cl}). Subsequently, we offer a comprehensive discussion on CL in~\cref{sec:CL}, highlighting its distinctions from other learning paradigms, addressing challenges, and reviewing the most recent and notable work in the sector. In~\cref{sec:RL}, we delve into RL by examining breakdowns and sub-segments, particularly focusing on learning environments and algorithms.

\subsection{Brain-inspired mechanisms for Continual Learning}
\label{sec:brain_cl}

Biological systems have showcased the inherent ability to continually learn by constant interactions with the surrounding environment. They seamlessly acquire experience, consolidate it with existing knowledge, and retain and reuse it as needed, all while interacting with dynamic environments regardless of the complexity of the task~\cite {sarfraz2023study}. This allows these systems to quickly adapt to unseen scenarios and learn new concepts efficiently despite learning from minimal demonstrations/interactions. 
Novel experiences contribute to improved performance and rarely cause forgetting of prior learnings~\citep{french1999catastrophic}. Consider driving a vehicle, a task that involves enhanced decision-making skills. In this scenario, a continuous improvement in the cognitive abilities of humans occurs over time, up until the ageing process begins.

\begin{figure*}[ht]
\centering
\begin{tabular}{{c@{ } c@{ } c@{ } }}
     {\includegraphics[width=0.5\linewidth, valign=t]{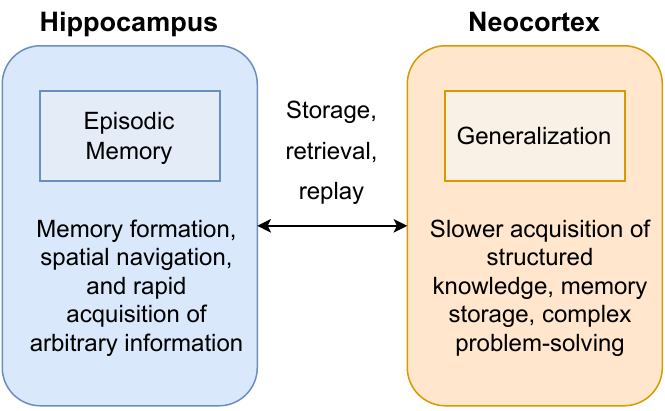}}&
     \hspace{1cm} &
    {\includegraphics[width=0.5\linewidth, trim={11cm 0cm 0cm 0.5cm},clip, valign=t]{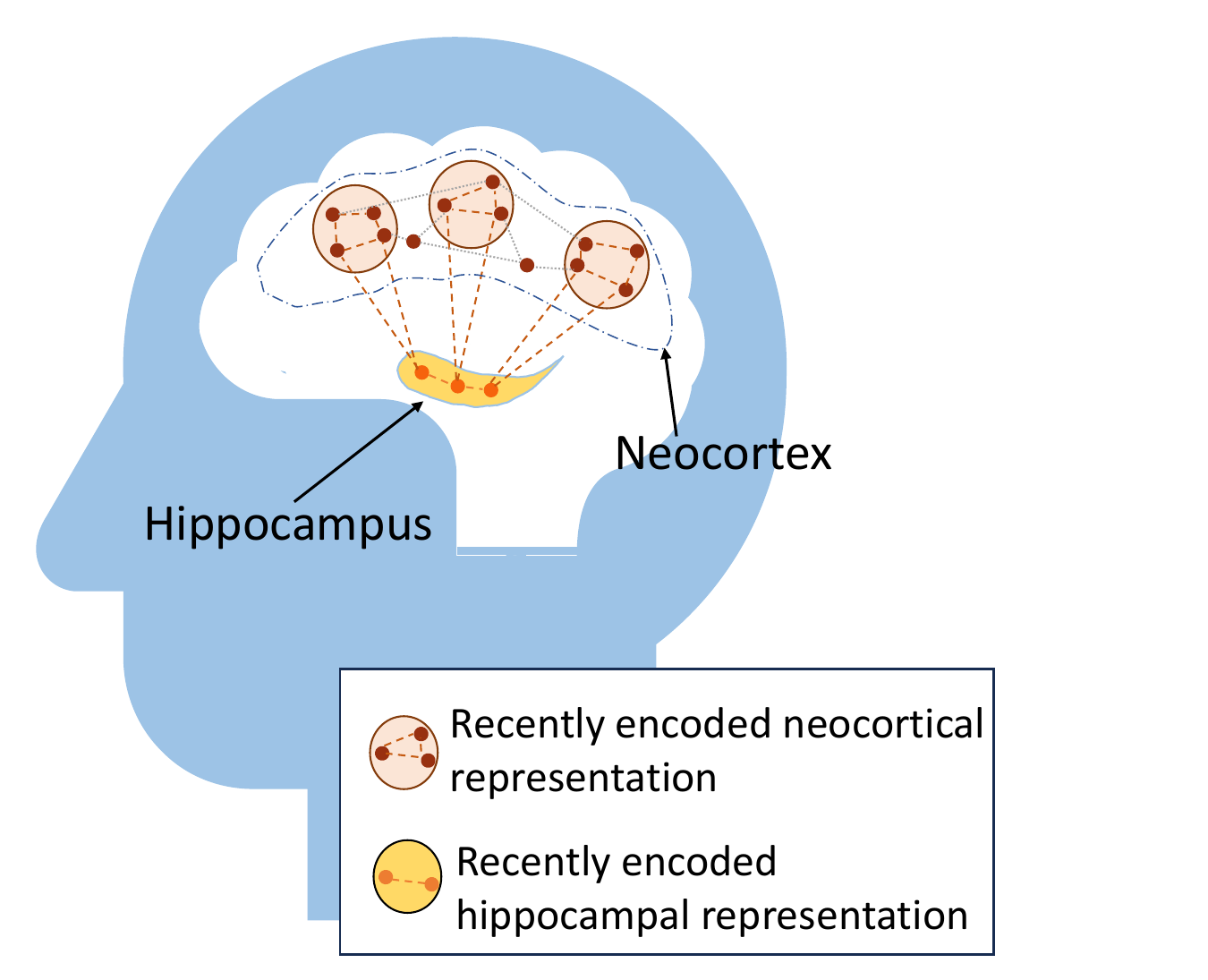}}\\
    (a) & & (b)\\
  \end{tabular}
\caption{The learning schema within the brain. (a) Complementary Learning System (CLS) theory from~\cite{parisi2019continual} (b) Memory consolidation within the brain inspired by~\cite{klinzing2019mechanisms}}
\label{fig:hippocampus}
\end{figure*}

The mammalian brain follows different strategies to maintain stability\footnote{Maintaining knowledge already gathered without forgetting.} across learned experiences and enhance plasticity\footnote{Ability to successfully incorporate new knowledge.}~\citep{mateos2019impact} to gain new knowledge efficiently while adapting to unseen situations.
\textbf{Episodic replay}~\citep{tulving2002episodic} involves the mechanism of replaying recent neuronal patterns that occurred during wake time to be repeated during rest/sleep time. This consolidates memory, overcomes forgetting, and initiates CL by transferring the contents from the short-term memory in the hippocampus to the long-term memory in the neocortex. The Complementary Learning Systems (CLS) theory~\citep{mcclelland1995there, kumaran2016learning}, shown in \Cref{fig:hippocampus}(a), states that the hippocampus is responsible for short-term adaptations to rapidly learn new information using a high learning rate whereas the neocortex employs a slower rating rate to learn general characteristics to be retained in the long run~\citep{parisi2019continual}. \Cref{fig:hippocampus}(b) shows how the hippocampus and neocortex are connected to transfer information within the brain.

\paragraph{Metaplasticity}~\citep{abraham1996metaplasticity}, also called the ``plasticity of the synaptic plasticity", refers to the strengthening or weakening of synaptic connections based on the significance of gained information. This process is essential for knowledge retention in the brain. The modification of the connection strength can be dependent on internal biochemical states, past synaptic modifications, and the latest neural activity. In the event of modifying the neural structure to accommodate new information, the brain reorganizes itself by forming new neurons~\citep{kempermann2002new, van2002functional, ge2007critical} in a process termed \textbf{Neurogenesis}. This takes place throughout life (more prominent in early development stages) while it is also capable of selectively forgetting or suppressing unwanted information to make room for new learning~\citep{hayes2021replay, kudithipudi2022biological}.

Though the standard artificial NNs in popular use today are inspired by the behavior and capabilities of the human brain, they are often designed to excel in tasks involving i.i.d. data. To imitate the brain in order to learn continually, the algorithmic structure of a simple artificial NN needs to be changed to better handle the non-stationary data distributions seen in CL settings. Without this, the standard NN faces catastrophic forgetting of prior learning. Just as diverse neurophysiological principles in the brain manage the stability-plasticity dilemma across multiple brain areas, it is crucial to identify corresponding machine learning algorithms to tackle similar challenges in CL within NNs.

\subsection{Continual Learning}
\label{sec:CL}

The most prevalent and common problem in CL is the task-incremental setting, wherein the training data ($\vx_{i}, \vy_{i}$) for each task $\task_{i}$ (with a total number of $\numtasks$ tasks) is presented sequentially from a series of tasks $\taskmath = \{\task_{1}, \task_{2}, \ldots, \task_{\numtasks}\}$. A task is characterized by a period during which the data distribution remains stationary (though this is not always the case), along with a specific objective function~\citep{lesort2020continual}. Each task consists of i.i.d. samples, where $\vx \in \data_{i}$ and $\vy \in \labels_{i}$ represent the input and the corresponding desired output for sample $i$, respectively. The mapping $f_i: \Theta \times \data_{i} \to \labels_{i}$ delineates the relationship between inputs and outputs for each task, where $\Theta$ denotes the parameter space of the model. In a CL context, as the learner receives task-specific training data along with the current task identifier $i$, its objective is to effectively address the current and previous tasks with minimal or no access to earlier data. This is achieved by minimizing the cost indicated by the loss function $\mathcal{L}(\theta)$ for task $t$.
\begin{align}
    \label{eqn:total_task_loss}
    \mathcal{L}_t(\theta) \triangleq \sum_{i=1}^t \E_{\vx,\vy \sim \data_{i} \times \labels_{i}} \Big[ \ell \big( f_i(\vx, \theta), \vy\big)\Big]\;.
\end{align}
Here, $\ell: \labels \times \labels \to \R_+$ is a loss function that measures the goodness of the prediction $\hat{\vy} = f(\vx, \theta)$ against the desired response $\vy$. Additionally, $\data = \bigcup_{i} \data_{i}$ and $\mathcal{Y} = \bigcup_{i} \labels_{i}$ represent the aggregate input and output spaces, respectively. The objective of \Cref{eqn:total_task_loss} is to identify the model parameters $\theta$ that minimize the loss across all tasks encountered up to time $t$.

\begin{figure}
    \centering
    \includegraphics[width=1\linewidth, trim={0cm 0cm 0cm 0cm},clip]{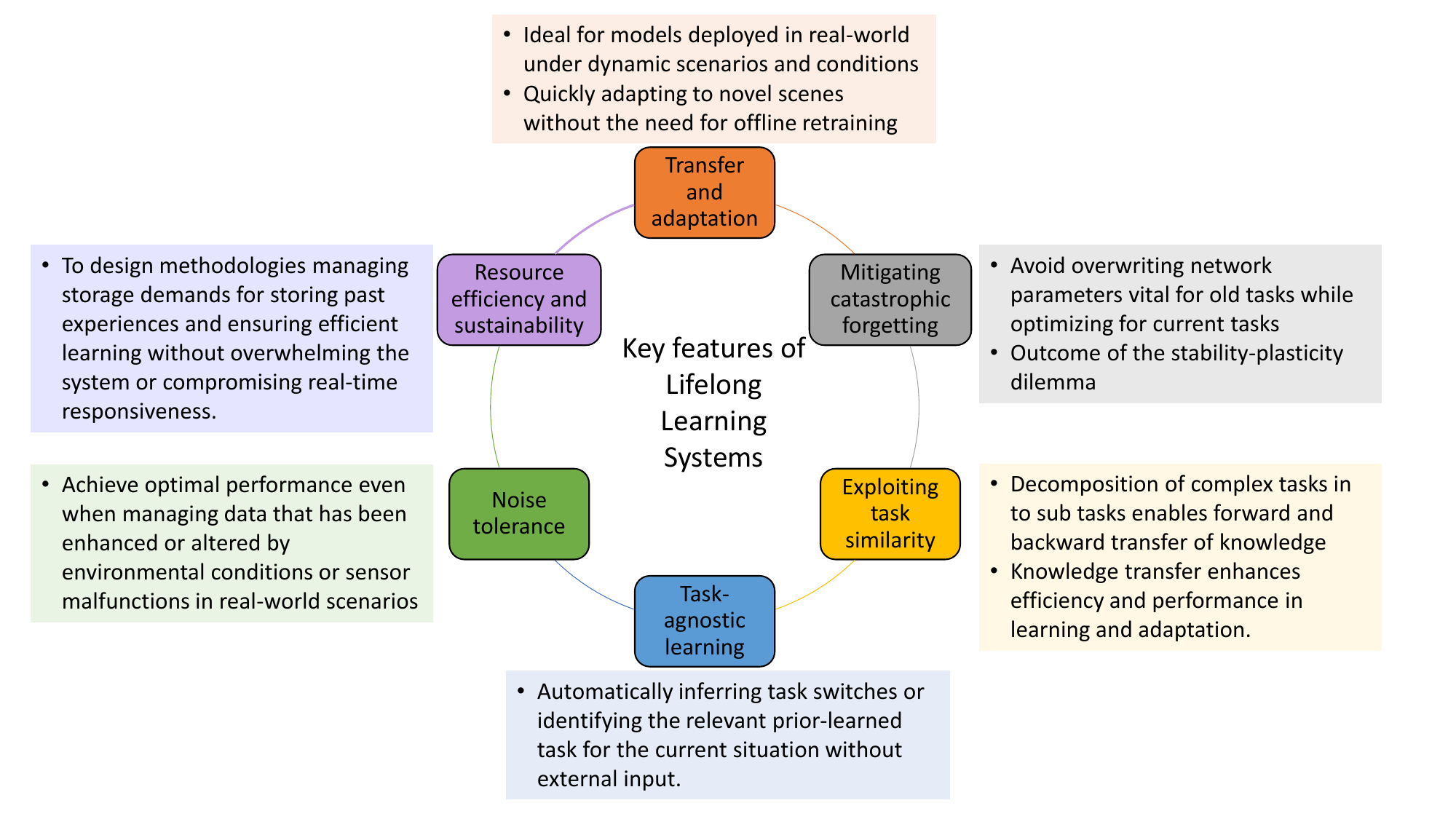}
    \caption{Key features/Desiderata of Lifelong Learning Systems~\citep{kudithipudi2022biological}.}
    \label{fig:lll_features}
\end{figure}

\paragraph{\textbf{Why do AI agents need to learn continually?}}
Solving single-task problems has enabled impressive milestones to be reached with respect to performance, processing speed, memory, and storage consumption~\citep{krizhevsky2012imagenet, vaswani2017attention, devlin2018bert, radford2021learning}. Isolated learning involves no further acquisition of new knowledge post deployment, nor the update of existing knowledge. To achieve genuine intelligence in AI systems, enabling them to learn, generalize, and apply knowledge from dynamic data streams sequentially, akin to human cognition, it is imperative to develop algorithms and neural architectures that are well-suited to these requirements. Hence, CL in deep NNs emerged as a line of study inspired by biological concepts~\citep{hassabis2017neuroscience}. It is important to highlight that the dynamic nature of the world, characterized by non-stationarity, underscores the need for NNs that are adept at adapting and aligning with CL's demands. Take autonomous vehicles as an example. This does not merely require AI agents that navigate urban infrastructure accurately. They must also possess the capability to assimilate new information, such as identifying and avoiding novel obstacles (\eg roadworks), and learn to traverse unfamiliar landscapes where conventional road signs might be insufficient. CL is also known as Lifelong Learning since the learning procedure can persist for an extended period of time. \Cref{fig:lll_features} depicts a summary of the ideal key features that define a successful Lifelong Learning system addressing the problems mentioned above. These features can be considered as desiderata of CL. 

\paragraph{\textbf{How do Knowledge Representations aid Continual Learning?}}
\cite{chen2018lifelong} discuss various types of knowledge representations. In contemporary CL research, previous knowledge typically functions as a form of prior information, such as prior model parameters or prior probabilities~\citep{rusu2016progressive}, for the upcoming task. Shared latent parameters~\citep{ruvolo2013ella}, learned model parameters~\citep{kirkpatrick2017overcoming}, results from past tasks~\citep{li2017learning}, items from information extraction models~\citep{liu2016improving}, and past relevant data~\citep{rebuffi2017icarl, xu2018lifelong} are examples of knowledge incorporated into new task data through various techniques and applications.
\newline
\newline
Current knowledge representation schemes lack universal applicability, making it challenging to use them consistently across various algorithms or task types. Shared knowledge could be global (shared across tasks of a similar nature due to high correlation or homogeneous distribution) or local (unique task-specific knowledge to selectively choose from). Global knowledge is appropriate to approximate optimality on past and current tasks but challenging in cases of highly diverse tasks. In contrast, local knowledge focuses on enhancing current task performance by leveraging prior knowledge or improving prior task performance by considering it as the current task. These methods have the flexibility to selectively incorporate specific aspects of past knowledge into ongoing learning or choose not to integrate them at all. Global knowledge-based methods prove effective when optimizing across all tasks, encompassing both past and current tasks as seen in multi-task learning. However, their efficacy diminishes when confronted with highly diverse tasks or a large number of tasks.

% trim=left bottom right top

\paragraph{\textbf{Why Continual Learning agents forget?}}
The primary challenge in CL lies in addressing catastrophic forgetting~\citep{french1999catastrophic, jedlicka2022contributions}, an outcome rooted in the stability-plasticity dilemma~\citep{carpenter1987massively}. When a model is trained on a new task without any CL mechanism, the NN adjusts the existing optimal weights linked to prior tasks to perform effectively in the current task. This creates an inherent challenge known as the stability-plasticity dilemma, where maintaining a balance between integrating new knowledge and preserving consolidated knowledge from past tasks becomes crucial. When the latest task requires significantly different representations to the previously learned tasks, the changes (to the weights of the NN) to accommodate the new knowledge are detrimental to the existing representations. CL methodologies are specifically designed to address this issue, allowing for the continual adaptation to evolving non-stationary data distributions. 

\paragraph{\textbf{How to combat forgetting while learning continually?}}
One of the earliest methods employed to combat catastrophic forgetting is memory replay, which entails regular re-exposure to previously seen samples mixed with new samples from the most recent data distribution~\citep{robins1993catastrophic}. A major limitation of this approach is the increasing storage demand to preserve old samples as new tasks emerge. This necessitates larger working memories and sophisticated learning algorithms with extensive computational requirements. 
In sensitive applications (\eg, healthcare), privacy concerns may also emerge with the use of memory replay. In situations where the NN architecture cannot be easily scaled, novel algorithms are sought to safeguard existing knowledge from being overwritten by the latest experiences.~\cite{richardson2008critical} states 3 ways to overcome catastrophic interference in connectionist networks. (i) New knowledge to be handled via new neural resources; (ii) Representations to be non-overlapping in fixed systems; (iii) Simultaneously refreshing old knowledge with the introduction of new information allows interleaving old and new knowledge within distributed representations over the same resource. As mentioned in \Cref{sec:brain_cl}, the mammalian brain exhibits sophisticated processes that facilitate seamless continual learning, a quality not fully replicated by conventional neural networks~\citep{ellis2021dreamcoder} paired with continual learning algorithms.

% \Cref{sec:cl_algo_breakdown} 
In the next part, we discuss the existing approaches for CL that, to an extent, address the problems highlighted in this section to better adapt networks in dynamic environments. In addition to addressing challenges in vision, CL has demonstrated success and improvements in various domains including robotics~\citep{lesort2020continual}, driving~\citep{shaheen2022continual}, medical~\citep{lee2020clinical, verma2023privacy}, anomaly detection~\citep{doshi2020continual, pezze2022continual} and more.

\subsubsection{Related Learning Paradigms}

\begin{figure}[ht]
    \centering
    \includegraphics[width=1\linewidth]{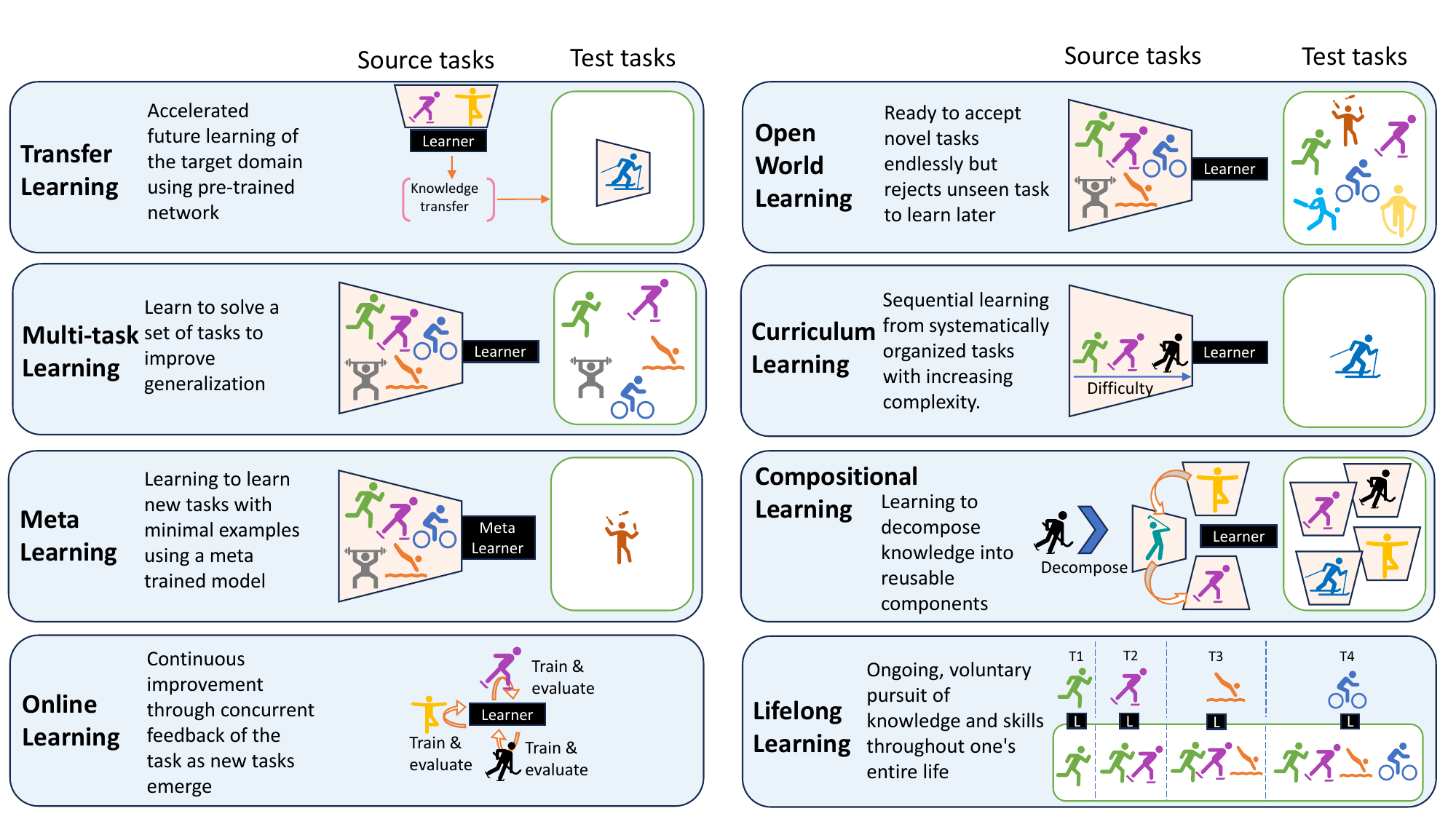}
    \caption{Diverse learning paradigms in contrast to CL. Each human figure represents a task. \textbf{Transfer Learning} - building future knowledge on past knowledge, \textbf{Multi-task Learning}- learning to balance multiple objectives, \textbf{Meta Learning} - learning to learn, \textbf{Online Learning} - adapting to the current task in real-time, \textbf{Open World Learning} - learning to embrace unknown tasks, \textbf{Curriculum Learning} - gradually learn to master skills of increasing complexity, \textbf{Compositional Learning} - learning to breakdown and piece knowledge together, \textbf{Lifelong Learning} - continuous knowledge growth journey.}
    \label{fig:other_algos}
\end{figure}

Learning can commence through different approaches, influenced by the accessibility and availability of data and the methodologies that enhance performance excellence in diverse settings. While CL can be seen as our grand challenge, there are many other closely related learning settings that align or can be considered steps towards more general sequential learning agents capable of operating in non-stationary environments. Below in \Cref{fig:other_algos} we provide a concise summary on this with examples to better differentiate CL.

Below, we briefly discuss and differentiate these closely related learning methods in comparison to CL. 
\paragraph{Transfer learning} enables agents to overcome the limitation of having to retrain statistical models from scratch each time the target data distribution changes. This is achieved by transferring knowledge (re-purpose learnt features) from already trained (pre-trained) models to target models. 
Transfer learning is preferred in scenarios where target data is limited due to unavailability or being outdated, as it is time and cost efficient and adaptable. The transfer technique used depends on what knowledge is to be transferred, how, and when, in relation to the source and target domains involved~\citep{pan2009survey, yosinski2014transferable}. The approach is motivated by the idea that individuals can smartly leverage previously acquired knowledge to solve new problems more quickly or effectively. Widely used methods include domain adaptation, multi-task learning, one-shot learning, zero-shot learning, knowledge distillation. We direct the user to follow the work from~\cite{pan2009survey, yosinski2014transferable} and~\cite{zhuang2020comprehensive} for a comprehensive overview of these techniques and their applications in various fields. Unfortunately, it is often the case that knowledge transfer results in forgetting, something explicitly CL aims to avoid. Applications of transfer learning methods include healthcare~\citep{shin2016deep, maqsood2019transfer}, transport~\citep{di2017cross} and natural language~\citep{devlin2018bert} contexts.
\paragraph{Multi-task learning} is a type of transfer learning methodology where a single model is trained to learn from multiple tasks simultaneously, leveraging the shared information between related tasks. This strategy helps improve data efficiency by extracting common representations, allowing the model to quickly adapt to new, unseen but related~\citep{kalashnikov2021mt} or out of distribution setups, particularly when paired with online learning~\citep{asenov2019vid2param, albuquerque2020improving} in the future~\citep{crawshaw2020multi, zhang2021survey}. Consequently, multi-task learning can enhance learning strategies that are both data and computationally efficient, reducing the need for extensive retraining when faced with novel challenges. However, it must also address potential conflicts where different tasks within the same group have contrasting requirements, which could impact individual task performance. This issue is known as \textit{negative transfer} or \textit{destructive interference}~\citep{standley2020tasks, fifty2021efficiently}.
% \paragraph{Meta-learning} introduces methodologies to adapt or generalize a model to tasks unseen during training, using minimal samples during test time. It improves the existing learning algorithm by exposing it to a multitude of related experiences/tasks during training in contrast to conventional learning methods that rely on multiple data instances~\citep{hospedales2021meta}. This aids to acquire generalizable knowledge for future tasks that may involve scenarios with limited data or compute resources~\citep{schmidhuber1987evolutionary, thrun1998learning}.
% MAML (Model-Agnostic Meta-Learning)~\citep{finn2017model} is a method that can be combined with many diverse architectures to learn new tasks with maximum performance by performing minimal number of gradient updates on the meta-learnt model. Reptile~\citep{nichol2018reptile} is another meta-learning algorithm (first-order version of MAML) that performs few-shot learning in scenarios with minimal training samples using SGD to locate optimal initialization parameters. Unfortunately, finetuning or downstream task improvements still leads to catastrophic forgetting of knowledge gained for prior tasks in this setting. In contrast to this, CL addresses to overcome this forgetting problem and enable models to maintain knowledge over multiple tasks as the data distribution changes.
\paragraph{Meta-learning} enhances a model's ability to adapt to new, unseen tasks using minimal samples, by exposing it to a wide range of related tasks during training. Unlike traditional learning, which depends on extensive data~\citep{hospedales2021meta}, meta-learning helps acquire generalizable knowledge for future tasks, even in scenarios with limited data or compute resources~\citep{schmidhuber1987evolutionary, thrun1998learning}. Techniques like MAML (Model-Agnostic Meta-Learning)~\citep{finn2017model} and Reptile~\citep{nichol2018reptile} (first-order version of MAML) enable quick adaptation to new tasks with minimal gradient updates. Unfortunately, these methods can lead to catastrophic forgetting, where the model loses knowledge of prior tasks when fine-tuning for new ones. CL on the other hand seeks to address this by enabling models to retain knowledge across multiple tasks as the data distribution evolves.
\paragraph{Online learning} in contrast to traditional batch or offline machine learning approaches, trains a model in real time as data arrives sequentially. With each new data point, the model is updated to improve future predictions, offering efficient and scalable solutions for large-scale, real-world problems~\citep{hoi2021online}. However, due to its incremental parameter update mechanism, online learning often struggles to retain prior knowledge, leading to CF, a challenge that CL aims to overcome.
% \paragraph{Online learning} in contrast to traditional batch or offline machine learning approaches, trains a model in real time as data arrives sequentially. For each step the online learner updates the best predictor for future data, thereby providing efficient and scalable solutions for large-scale real-world problems~\citep{hoi2021online}. Due to the incremental parameter update (adaptation) mechanism, the online learning agent fails to retain prior knowledge leading to CF which CL aims to address.
% \paragraph{Open world learning} prepares the model to realize unseen and unknown data samples at any point in time to be rejected and to be learnt at a later time period~\citep{bendale2016towards}. NNs tend to choose a class label with high confidence scores for unknown and unrelated images in a closed set problem, where as open world learning tries to overcome this error. Although CL currently focuses on evaluating problems in a closed world setup, it is vital to enable them with knowledge to solve problems in an open world setting to successfully meet challenges when deployed in real-world approaches.
\paragraph{Open world learning} equips models to identify and reject unseen or unknown data samples, allowing them to be learned at a later time~\citep{bendale2016towards, mundt2023wholistic}. Unlike traditional NNs, which often assign high-confidence class labels to unfamiliar or unrelated images in closed-set scenarios, open world learning aims to mitigate this issue. While CL typically evaluates models in a closed world context, enabling them to handle open world challenges is crucial for effective deployment in real-world applications.
% is more often seen in situations where curiosity, intrinsic motivation or RL with empowerment. 
\paragraph{Curriculum learning} involves training NN models by starting with simple tasks and gradually progressing to more complex ones, rather than using randomly selected examples like many state-of-the-art models. This approach can lead to improved performance and faster convergence during training~\citep{soviany2022curriculum, minelli2023towards}. Curriculum learning reflects the strategy followed by humans where concepts are built up in stages of increasing complexity for meaningful understanding. Hence, this could be considered as a specific instance of CL where the gradual learning process inherently limits CF. An important contribution in this field \wrt RL is by~\cite{rudin2022learning}, where robots are trained to navigate various terrains with different difficulty levels. The proposed curriculum significantly reduces training time. 
\paragraph{Compositional learning} emulates the human brain's decision-making by using a collection of specialized NNs to handle different aspects of a task. Like brain components that work together (senses, memory, perception \etc), these NNs collaborate to achieve a desired outcome by breaking down complex tasks into manageable parts. This approach has been effectively demonstrated in RL through methods like the options framework~\citep{sutton1999between, konidaris2009skill}, skill trees~\citep{konidaris2012robot}, and hierarchical RL~\citep{florensa2017stochastic}, enabling more efficient problem-solving and adaptability.

\subsubsection{Breakdown of CL algorithms}
\label{sec:cl_algo_breakdown}

\begin{figure}
    \centering
    \includegraphics[width=1\linewidth, trim={0cm 0cm 0cm 0cm},clip]{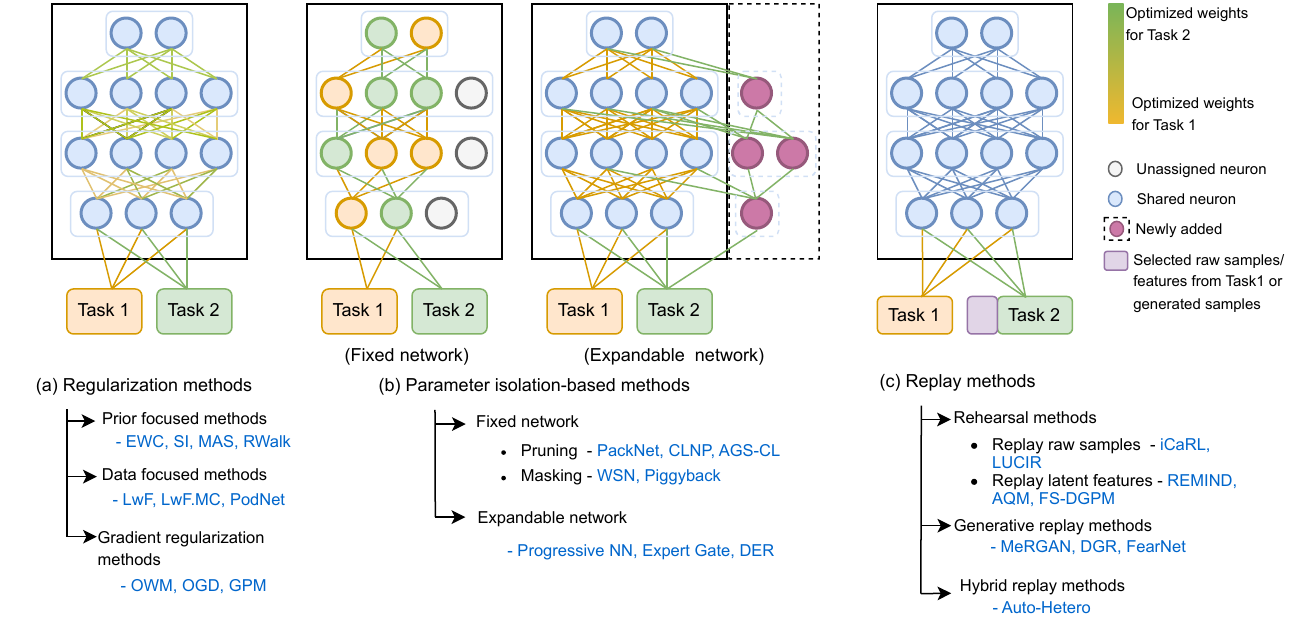}
    \caption{Classification of CL methodologies with their sub-categories. The blue text represents specific examples under each category}
    \label{fig:cl_methods}
\end{figure}

In this section, we will examine the approaches employed to address catastrophic forgetting resulting from the stability-plasticity dilemma in CL settings. \Cref{fig:cl_methods} provides an overview of how NNs are used to handle different CL methods. The figure also illustrates subcategories within regularisation-based, parameter isolation-based, and replay-based methods. 

\paragraph{Regularization approaches}
\label{para_regulaz_CL}

These algorithms incorporate a penalty term into the loss function, regulating the model's parameter updates~\citep{peters2010relative, kirkpatrick2017overcoming, zenke2017continual, aljundi2018memory, ritter2018online, chaudhry2018riemannian, aljundi2019task, zhuo2023continual}. This reduces the risk of overfitting to the most recent task and safeguards crucial weights associated with prior tasks, preserving optimal performance. Additionally, by not storing raw samples, this approach prioritizes data privacy and alleviates the need for additional storage.

The constrained weight updates limit plasticity on parameters crucial for earlier tasks, making the backward transfer of knowledge challenging~\citep{khetarpal2022towards}. In \textbf{prior-focused} regularisation methods, the penalty is applied selectively to control the weight change of parameters to preserve knowledge of prior tasks. This change in the parameter is proportional to the importance factor relevant to solving previous tasks. The pseudo-code for this particular approach of CL is depicted in \Cref{alg:prior-focused}. 
\newline
In the algorithms defined below $\delta$ is a metric of similarity between $\vtheta$ and $\vtheta^\ast_{\texttt{prev}}$ wrt to $\mathcal{I}_g$. 
 An example is $\delta\big(\mathcal{I}_{g}, \vtheta, \vtheta^\ast_{\texttt{prev}}\big) = \sum_{i=1}^n \mathcal{I}_{g}[i] \big|\vtheta[i]-\vtheta^\ast_{\texttt{prev}}[i]\big|^2$. The task loss is defined as the Cross-Entropy loss. 
\newline
\newline
Elastic Weight Consolidation (EWC)~\citep{kirkpatrick2017overcoming} works in a similar manner to how task-specific synaptic consolidation takes place in the human brain. It preserves knowledge from previous tasks by adding a quadratic penalty term to the loss function that discourages changes/plasticity to weights vital to maintaining performance on the earlier learned tasks. At convergence, the posterior distribution $p(\theta|D_{A})$ encapsulates vital information on important parameters for Task A. In order to store this information while optimizing for the next task, Task B, the Laplace approximation technique has been used as a guide to assume the approximation of the posterior as a Gaussian distribution. The optimal parameters $\theta_{A}^{\ast}$ and the diagonal of the Fisher Information Matrix (FIM)($F_{i,i}$) are used here. The FIM is defined as: 
% \AZ{new info on FI follows}
% Fisher Information (FI) is a measure of unit used to access the amount of information a random variable ($X$) carries about an unknown parameter ($\theta$) which the probability of $X$ is dependent upon. The probability density function (PDF) of $X$ given by $f(X; \theta)$, describes the outcome of $X$ for a given value of $\theta$. FI is defined as the variance of the score function \cref{eq:score} given as the square of the partial derivative \wrt $\theta$ of the natural logarithm of the likelihood function \cref{eq:fish}. 
% \begin{equation}
%     % I(\theta) = Var\big(\frac{\partial }{\partial \theta}ln(f(X; \theta)\big)
%     \texttt{Score} = \frac{\partial }{\partial \theta} \log(f(X; \theta)
%     \label{eq:score}
% \end{equation}
\begin{equation}
    F_{i,j} = \mathbb{E}\left[ \left(\frac{\partial}{\partial \vtheta_i} \log f\big(\vx;\vtheta\big)\right)
    \left(\frac{\partial}{\partial \theta_j} \log f\big(\vx;\theta\big) \right)\right]
    % \mathbb{E}\big[\big(\frac{\partial }{\partial \theta} \log(f(X; \theta)\big)\big]
    \label{eq:fish}
\end{equation}
%\MH{is it square ort something like $JJ^\top$, which ref are you using here?}\AZ{\href{https://en.wikipedia.org/wiki/Fisher_information#cite_note-7}{wiki}}
% \MH{define it mathematically and provide an intuitive explanation (amount of information X carries about $\theta$. Check the \href{https://en.wikipedia.org/wiki/Fisher_information}{Wiki}}
%
%FIM is a generalization of the FI in the case where $N$ parameters are to be estimated such that, $\theta  = (\theta_1, \theta_2, \ldots, \theta_N)$. This resultant FIM takes the form of a $N \times N$ matrix that contains the FI for all possible pairs of parameters. While this signifies the importance/strength of each weight in relation to the prior task, 
The diagonal elements of the FIM can be understood as the curvature of the loss function \wrt to each weight parameter.
% by computing the squared expectation of the gradients of the loss function \wrt each weight. 
This reflects the sensitivity of the network to changes in individual weights and helps determine the regularization strength to preserve knowledge. In~\Cref{eq:2}, for the current parameter $i$, a close to zero $F_{i}$ value indicates the minimal contribution of parameter $\theta_{i}$ to Task A performance. Hence, this parameter faces less constraint to adapt to Task B compared to parameters associated with a higher value for $F_{i}$. Conversely, such parameters would be imposed with a stronger constraint to keep it unchanged to maintain its contribution to Task A. The loss function to minimize is given by, 

\begin{equation}
    \mathcal{L(\theta)} = \mathcal{L_{B}(\theta)} + \sum_{i} \frac{\lambda}{2} F_{i,i}(\theta_{i} - \theta^{*}_{A,i})^2
    \label{eq:2}
\end{equation}
where $\mathcal{L_{B}(\theta)}$ is the task B loss, $\lambda$ the regularization coefficient balances the trade-off between adapting to the new task while retaining knowledge on the old task and $\theta_{i}$ the current parameter weight.

Motivated by the complex process undertaken by synapses in the brain that affects plasticity,~\cite{zenke2017continual} presented Synaptic Intelligence (SI), an equivalent in artificial NNs. The importance of each synapse symbolizes its contribution towards change in the total loss. These are maintained as an online estimate while training so that memory consolidation takes place as soon as the task changes, meaning changes to these synapses/weights are penalized. MAS~\citep{aljundi2018memory} introduced an unsupervised and online approach to compute parameter importance for regulating weight changes during new task learning. A combination of SI and EWC was used to develop RWalk by~\cite{chaudhry2018riemannian}. The Euclidean space-based sensitivity measure in SI was replaced by a KL divergence measure between output distributions while EWC was updated to an online form to function with efficient memory. Other work in this context includes~\cite{lee2017overcoming, park2019continual, ebrahimi2019uncertainty}. BLIP~\citep{shi2021continual} employed this technique to retain information gain in model parameters by updating them at the bit-level.

\begin{algorithm}
    \SetAlgoLined
    \footnotesize
    \caption{Pseudo-code for Prior-focused regularisation in CL}\label{alg:prior-focused}

        \SetAlgoLined
        \KwInput{New task training data $\mathcal{D} = \{(\vx_j,\vy_j)\}_{j=1}^{m}$,\\ 
        previously learned parameters $\vtheta^\ast_{\texttt{prev}} \in \mathbb{R}^n$, \\
        importance/sensitivity parameters $\mathcal{I}_g \in \mathbb{R}^n$, \\
        regularization coefficient $\lambda \in \mathbb{R}$, number of epochs $ep$.}        
        \KwOutput{New parameters $\vtheta^\ast \in \mathbb{R}^n$, \\
        updated importance/sensitivity parameters $\mathcal{I}_g$.}
        $\vtheta \gets \vtheta^\ast_{\texttt{prev}}$\;
        \For {epoch $e = 1$ \KwTo $ep$} {
            $\hat{y} \gets f(\vx;\vtheta),\quad \forall \vx \in \mathcal{D}$; obtain the output of the model for data in the current task\;             
            $\mathcal{L}(\vtheta) \leftarrow \mathbb{E}_{(\vx,\vy) \sim \mathcal{D}}\texttt{Loss}(\vy,\hat{\vy}; \vtheta)$ ; Compute the task loss\;
            % $\mathcal{L}_{Total} \leftarrow \min(\mathcal{L}_{i} + \sum \lambda * (\mathcal{I}_{z}(\theta_{i},\theta_{i-1})))$ 
            $\vtheta \leftarrow \argmin_\vtheta \Big(\mathcal{L}(\vtheta) + \lambda \delta\big(\mathcal{I}_{g}, \vtheta, \vtheta^\ast_{\texttt{prev}}\big)\Big)$\;
        }            
        Update $\mathcal{I}_g$ based on its previous values and current data\;
\end{algorithm}

% \MH{ALso, you need to define Loss, delta and etc. Here is how I define $\delta$}
% \textcolor{red}{
%  $\delta$ is a metric of similarity between $\vtheta$ and $\vtheta^\ast_{\texttt{prev}}$ wrt to $\mathcal{I}_g$. 
%  An example is $\delta\big(\mathcal{I}_{g}, \vtheta, \vtheta^\ast_{\texttt{prev}}\big) = \sum_{i=1}^n \mathcal{I}_{g}[i] \big|\vtheta[i]-\vtheta^\ast_{\texttt{prev}}[i]\big|^2$
% }

On the other hand, \textbf{data-focused} regularisation techniques focus on knowledge distillation-based learning. A model trained on past tasks (teacher) shares its knowledge with a new model (student) to adapt continually and mitigate CF. The soft target is reserved using a distillation loss that is added to the objective function~\citep{li2017learning, rannen2017encoder, dhar2019learning, douillard2020podnet}. \Cref{alg:data-focused} displays the pseudo-code for data-focused CL algorithms. LwF~\citep{li2017learning} and LwF. MC~\citep{rebuffi2017icarl} uses the outputs from the model trained on previous tasks as soft labels for the current task. 

Although data-focused methods are rarely used in isolation, they have been used in combination with other CL methodologies to reduce CF as discussed in the upcoming sections. 

% \AZ{https://arxiv.org/pdf/1705.08395.pdf}
% \AZ{has ewc equation}

Addressing challenges in scenarios where the model size is constrained (preventing expansion), and the agent lacks access to past data via replay-based methods is particularly intriguing, as it pertains to numerous practical problems. An alternative path is projection-based approaches in the parameter space. Some work on this line of research includes~\cite{zeng2019continual, farajtabar2020orthogonal, chaudhry2020continual, saha2021gradient}. To overcome the constraints pointed out earlier, OWM~\citep{zeng2019continual} fuses an orthogonal weight modification method with a context-dependant processing module to enable CL of context-specific mappings for a classification use case. The method needs improvement with scalability and efficient computations.~\cite{farajtabar2020orthogonal} presents OGD, a novel technique where, the gradient directions corresponding to past tasks saved in memory are used to update the gradients for the new task in an orthogonal direction (to the gradient subspace representing past tasks). This introduces minimal interference to knowledge from past tasks, hence minimizing CF.

\begin{algorithm}[h]
    \SetAlgoLined
    \footnotesize
    \caption{Pseudo-code for {Data-focused} regularisation in CL}\label{alg:data-focused}
        \SetAlgoLined
        \KwInput{New task training data $\mathcal{D} = \{(\vx_j,\vy_j)\}_{j=1}^{m}$,\\
        teacher model parameters $\vtheta^\ast_{\texttt{T}} \in \mathbb{R}^n$, \\
        student model parameters $\vtheta_{\texttt{S}} \in \mathbb{R}^n$, \\
        regularization coefficient $\lambda \in \mathbb{R}$, number of epochs $ep$.}        
        \KwOutput{New student parameters $\vtheta^\ast_{\texttt{S}} \in \mathbb{R}^n$.}
        $\hat{y}_{\texttt{T}} \gets f(\vx;\vtheta^\ast_{\texttt{T}}),\quad \forall \vx \in \mathcal{D}$; obtain the output of the teacher model for data in [1, n] tasks\;
        
        % $\vtheta \gets \vtheta^\ast_{\texttt{prev}}$\;
        \For {epoch $e = 1$ \KwTo $ep$} {
            $\hat{y}_{\texttt{S}} \gets f(\vx;\vtheta_{\texttt{S}}),\quad \forall \vx \in \mathcal{D}$; obtain the output of the model for data in the current task\;             
            $\mathcal{L}(\vtheta_{\texttt{S}}) \leftarrow \mathbb{E}_{(\vx,\vy) \sim \mathcal{D}}\texttt{Loss}(\vy,\hat{\vy}_{\texttt{S}}; \vtheta_{\texttt{S}})$ ; Compute the task loss\;
            $\mathcal{L}_{\texttt{Dis}}(\vtheta_{\texttt{S}}) \leftarrow \texttt{Loss}(\hat{\vy}_{\texttt{S}}, \hat{\vy}_{\texttt{T}}; \vtheta_{\texttt{S}})$ ; Compute the distillation loss\;
            $\vtheta^\ast_{\texttt{S}} \leftarrow \argmin_{\vtheta_{\texttt{S}}} \Big(\mathcal{L}(\vtheta_{\texttt{S}}) + \lambda \mathcal{L}_{\texttt{Dis}}(\vtheta_{\texttt{S}})\Big)$\;
        }            
\end{algorithm}

\paragraph{Parameter isolation methods (Architectural approaches)}
    The studies carried out under this segment often utilize the sparsity of NNs. Sparsity is the excessive presence of zero or near-zero weights in a NN that, regardless of its presence, may not actively contribute to the current learning process. Many approaches in this sector have tried to take advantage of the over-parameterization of NNs~\citep{neyshabur2018towards}. A detailed review on sparsity can be found in~\cite{gale2019state}. Pruning~\citep{fernando2017pathnet, mallya2018packnet, wang2020learn} and masking~\citep{serra2018overcoming, mallya2018piggyback, koster2022signing, konishi2023parameter, ke2023sub} techniques strategically select and set certain weights to zero within \textbf{fixed architectures}, effectively isolating parameters associated with each task. This allows parts of the NN to be activated while others are deactivated, providing optimal task-wise sub-networks and parameter separation. This is beneficial in overcoming CF (due to non-interfering neural connections), accommodating new tasks and optimizing computational resource utilization. When expanding network capacity is not a constraint (\textbf{dynamic architectures}), new neurons are incorporated into the architecture allowing additional room to learn new tasks~\citep{rusu2016progressive, aljundi2017expert, yoon2017lifelong, xu2018reinforced, ebrahimi2020adversarial, yan2021dynamically} or new connections are added while reusing old components allowing forward knowledge transfer~\citep{sokar2022avoiding}. Limitations of this approach include scalability issues when learning a long sequence of tasks and the inefficiency introduced by the overhead of searching through architectures during inference. Respective pseudo-codes are provided in~\Cref{alg:fixed} and~\Cref{alg:dynamic}.

If the network architecture is kept fixed, the algorithm focuses on the optimal parameter allocation for each task to maintain sufficient capacity for future tasks by creating subnetworks. The mission of agents in PathNet~\citep{fernando2017pathnet} is to find the best adaptive pathways within the network to be re-used for new tasks using a genetic algorithm~\citep{fernando2011evolvable}. The unused weights get reinitialized when a new task is encountered, leaving the used weights frozen. This decision is based on experimental findings indicating that the transfer performance did not surpass that achieved through fine-tuning unless re-initialization was applied. Another adaptive sparse learning mechanism has been performed to learn continually in SpaceNet by~\cite{sokar2021spacenet}. Activation-based sparsity pruning (sparsity in the number of neurons used) has been tested by CLNP~\citep{golkar2019continual} and AGS-CL~\citep{jung2020continual} as a form of parameter-isolation method for CL.

PackNet~\citep{mallya2018packnet} employed a weight-based pruning technique to free up parameters across all layers, creating room to learn future tasks. Post pruning, the optimal parameters for each task are frozen ensuring to maintain accuracy for the learnt task. This pruning technique generates task-specific parameter masks, isolating the parameters and sustaining performance across all tasks with minimum storage costs per task. The specific mask needs to be provided during inference since there is no mechanism to identify the task being tested. A similar attempt inspired by the Lottery Ticket Hypothesis~\citep{frankle2018lottery} combined with weight score-based binary masking method was carried out by~\cite{kang2022forget} (WSN) generating subnetworks for each task. This forget-free technique ensured minimal CF and forward transfer of knowledge by reusing frozen parameters with the help of the binary mask. The binary mask was compressed using Huffman encoding to reduce storage space. Other closely related works in masking include PiggyBack~\citep{mallya2018piggyback}, SupSup~\citep{wortsman2020supermasks} and~\cite{carreno2023adapting}.

\begin{algorithm}[h]
    \SetAlgoLined
    \footnotesize
    \caption{Pseudo-code for CL in fixed architecture}\label{alg:fixed}
    
        \SetAlgoLined
        \KwInput{New task training data $\mathcal{D} = \{(\vx_j,\vy_j)\}_{j=1}^{m}$,\\
        newly initialised parameters $\vtheta \in \mathbb{R}^n$, \\
        pruning percentage $p \in (0,1)$, binary mask $M \in \{0,1\}^n$,  number of epochs $ep$.}        
        \KwOutput{New parameters $\vtheta^\ast \in \mathbb{R}^n$.}
        \For {epoch $e = 1$ \KwTo $ep$} {
            \uIf{Prune}{
                $\hat{y} \gets f(\vx;\vtheta),\quad \forall \vx \in \mathcal{D}$; obtain the output of the model for data in the current task\;
                $\mathcal{L}(\vtheta) \leftarrow \mathbb{E}_{(\vx,\vy) \sim \mathcal{D}}\texttt{Loss}(\vy,\hat{\vy}; \vtheta)$ ; Compute the task loss\;
                $\vtheta^\prime \gets f(p; \vtheta)$ ; Prune the network\;
                $\hat{y}_{\texttt{p}} \gets f(\vx;\vtheta^\prime),\quad \forall \vx \in \mathcal{D}$; obtain the output of the pruned model for data in the current task\;
                $\mathcal{L}(\vtheta^\prime) \leftarrow \mathbb{E}_{(\vx,\vy) \sim \mathcal{D}}\texttt{Loss}(\vy,\hat{\vy}_{\texttt{p}}; \vtheta^\prime)$ ; Compute the loss due to pruning\;
                $\vtheta^\ast \leftarrow \argmin_\vtheta \Big(\mathcal{L}(\vtheta) + \mathcal{L}(\vtheta^\prime)\Big)$\;
            }
            \uElseIf{Mask}{
                $\vtheta^\prime \gets f(M; \vtheta)$ ; Masked network\;
                $\hat{y}_{\texttt{M}} \gets f(\vx;\vtheta^\prime),\quad \forall \vx \in \mathcal{D}$; obtain the output from the masked model for data in the current task\;
                $\mathcal{L}(\vtheta^\prime) \leftarrow \mathbb{E}_{(\vx,\vy) \sim \mathcal{D}}\texttt{Loss}(\vy,\hat{\vy}_{\texttt{M}}; \vtheta^\prime)$ ; Compute the task loss\;
                $\vtheta^\ast \leftarrow \argmin_\vtheta \Big(\mathcal{L}(\vtheta^\prime)\Big)$\;
            }
        }            
\end{algorithm}

Progressive Networks~\citep{rusu2016progressive} use a growing NN architecture to reduce intervention between parameters optimized for past tasks by freezing them while adding new parameters to acquire new knowledge. It uses lateral connections between layers that enable the incorporation of learnings from past tasks to current ones, promoting shorter learning times. This effectively promotes the forward transfer of knowledge. Network of Experts (Expert Gate)~\citep{aljundi2017expert} adds specialists as new tasks arrive while knowledge from relevant previous experts is transferred to this new expert. In this setup, autoencoders learn task representations and the relatedness between each task to select the most relevant expert to train the new expert on. During inference, a gating mechanism learned by the autoencoder, selectively activates and loads into memory the relevant expert trained for a similar task based on the lowest reconstruction error.

A dual-stage learning mechanism aiming to expand the base network has been employed by DER~\citep{yan2021dynamically}. During the representation learning stage, the learned representations are frozen and augmented with novel feature dimensions using a learnable feature extractor to improve stability. This involves a channel-level mask-based pruning strategy. In the second stage (classifier learning), a balanced finetuning method is used to retrain the classifier to overcome biases introduced by class imbalance. Results indicate positive backward and forward transfer of knowledge.

\begin{algorithm}[h]
    \SetAlgoLined
    \footnotesize
    \caption{Pseudo-code CL for dynamic architectures} \label{alg:dynamic}
    
        \SetAlgoLined
        \KwInput{New task training data $\mathcal{D}_i = \{(\vx_j^{(i)},\vy_j^{(i)})\}_{j=1}^{m_i}$,\\ 
        initial parameters $\vtheta \in \mathbb{R}^n$, \\
        expandable components (neurons, layers) $w$, \\
        number of epochs $ep$.}        
        \KwOutput{New parameters $\vtheta_{\text{aux}} \in \mathbb{R}^n$, updated model $\vtheta^\prime$.}
        \For {each task $i$} {
        \For {epoch $e = 1$ \KwTo $ep$} {
                $\hat{y} \gets f(\vx;\vtheta),\quad \forall \vx \in \mathcal{D}_{i}$; obtain the output of the model for data in the current task\;             
                $\mathcal{L}(\vtheta) \leftarrow \mathbb{E}_{(\vx,\vy) \sim \mathcal{D}_{i}}\texttt{Loss}(\vy,\hat{\vy}; \vtheta)$ ; Compute the task loss\;
                $\vtheta_{\text{aux}}^\ast \leftarrow \argmin_\vtheta \Big(\mathcal{L}(\vtheta)\Big)$\;
            }            
            $\vtheta_{\text{aux}}^\ast$ ; Freeze the model\;
            $\vtheta^\prime \gets f(w; \vtheta_{\text{aux}}^\ast)$ ; Expanded network\;
        }
        
\end{algorithm}

\paragraph{Replay-based approaches}
\label{lab:replay}

A replay of neural activities representing prior experiences in the hippocampus and neocortex is said to solidify the memory formation, consolidation, and retrieval of knowledge~\cite{nadasdy1999replay, walker2004sleep}. However, it is important to note that the brain does not save visual images; instead, it saves the underlying patterns, associations, and contextual information associated with those images~\citep{pylyshyn2003return}. A similar process has been adopted for CL in artificial NNs where \textbf{past experiences (raw images) are stored} in a buffer to be replayed while training on data from a new task~\citep{rebuffi2017icarl, chaudhry2018efficient, rolnick2019experience, hou2019learning, kim2020imbalanced, zhuo2023continual, liang2024loss}. Although replaying the entire set of observed data while learning each new task would mitigate CF, it is not a practical approach. This is due to the surging storage requirement and privacy concerns in some cases. Thus, only a selected portion of seen data could be set aside for retraining. Another set of works has tried to replicate the replay strategy of the brain by \textbf{rehearsing compressed representations}~\citep{hayes2020remind, caccia2020online, deng2021flattening, he2022online, saha2023online}. Here, the challenge lies in selecting the most appropriate NN layer to transfer representations from. We direct the reader to~\cite{hayes2021replay}, for a comprehensive study on replay for CL. \Cref{alg:replay} shows pseudo-code detailing the implementation of the replay mechanism.
\newline

% \MH{my version below}
\begin{algorithm}[h]
    \SetAlgoLined
    \footnotesize
    \caption{Pseudo-code for replay-based CL}\label{alg:replay}
    
    \KwInput{New task training data $\mathcal{D}_i = \{(\vx_j^{(i)},\vy_j^{(i)})\}_{j=1}^{m_i}$,\\ 
        initial parameters $\vtheta \in \mathbb{R}^n$, \\
        replay buffer $\mathcal{B}$, batch size $b \in \mathbb{N}$, \\
        number of epochs $ep$.}        
    \KwOutput{Updated parameters $\vtheta^\ast \in \mathbb{R}^n$.}
    
    \For {each task $i$} {
        \For {epoch $e = 1$ \KwTo $ep$} {
            \For {each mini-batch} {
                $\mathcal{D}_1 \gets \texttt{Sample}(\mathcal{D}_i, b)$\;
                $\mathcal{D}_2 \gets \texttt{Sample}(\mathcal{B}, b)$\;
                $\mathcal{D}_{batch} \gets \mathcal{D}_1 \cup \mathcal{D}_2$\;
                $\hat{\vy} \gets f(\vx; \vtheta)$ for $(\vx, \vy) \in \mathcal{D}_{batch}$\;
                $\mathcal{L}(\vtheta) \leftarrow \mathbb{E}_{(\vx,\vy) \sim \mathcal{D}_{batch}} \texttt{Loss}(\vy, \hat{\vy}; \vtheta)$\;
                $\vtheta \leftarrow \vtheta - \eta \nabla_\vtheta \mathcal{L}(\vtheta)$\;
            }
        }
        $\mathcal{B} \leftarrow \mathcal{B} \cup \texttt{Sample}(\mathcal{D}_i, k)$ \textit{(Add $k$ samples from the new task to the buffer)}\;
    }
\end{algorithm}

\begin{remark}
The replay buffer $\mathcal{B}$ in CL can be updated in various ways. A simple strategy is to update it by adding $k$ samples from the new task. This process may involve replacing the oldest or least useful samples to maintain a fixed buffer size, ensuring that the buffer does not grow indefinitely.
\end{remark}

A detailed investigation has been carried out by~\cite{merlin2022practical} to understand how various settings affect the efficient performance of different replay based methods in CL under four benchmarks. Design choices such as the type of replay buffer, memory capacity, weighting policies and data augmentation techniques have shown profound effects on replay based CL methods. Herd selection~\citep{welling2009herding}, entropy-based selection~\citep{chaudhry2018riemannian} and discriminative sampling~\citep{liu2020mnemonics} are some exemplar selection methods in use. High-quality sample selection mechanisms~\citep{nguyen2017variational, isele2018selective, aljundi2019gradient}, compressing raw image data into feature representations~\citep{hayes2020remind, pellegrini2020latent} and selective sample retrieval for better parameter updates~\citep{shim2021online} have been recent advancements to improve rehearsal methods at times bypassing the need to reserve raw samples.

iCaRL by~\cite{rebuffi2017icarl} manages to learn continually using a herding-based replay method in combination with a distillation technique. As a class incremental setting, the reserved samples closely approximate the class mean in its learned feature space. A nearest-mean-of-exemplars approach has been adapted for classification, requiring only a minimal number of exemplars. Feature representation learning is carried out using the exemplars and distillation of knowledge between different time points (instead of between different models) to mitigate further forgetting. In LUCIR~\citep{hou2019learning}, a unified multi-class classifier was introduced for incremental learning, addressing imbalanced data issues in replay-based CL. It includes a cosine normalization layer instead of softmax to handle magnitude bias, a less-forget constraint (loss) to retain previous learnings, and inter-class separation (a margin ranking loss) supported by samples reserved through herding.

A time scheduled replay mechanism has been studied by~\cite{klasson2022learn} based on the findings from human learning methods claiming learning followed by spaced repetition at certain time intervals~\citep{dempster1989spacing, willis2007review} improve memory retention. Here, they learn the time segment during which a certain task has to be replayed to consolidate it to the memory while adapting to the new task. This takes place under the assumption that all historical data is available due to cheap data storage but, the replay memory size is constrained to hold historical data only once. SER~\citep{zhuo2023continual} is a hybrid approach between experience replay and regularization. It incorporates backward and forward consistency losses, enabling the distillation of knowledge to overcome forgetting.

In contrast to general replay methods, \textbf{pseudo-rehearsal}~\citep{robins1995catastrophic} was introduced as an alternative method. It artificially generates pseudo-items resembling the prior population the model was trained on, surpassing the need to access the true prior population. This enables the model to maintain its knowledge by rehearsing using generated samples on past tasks while learning new tasks. \Cref{alg:gen-replay} outlines the respective pseudo-code implementation for pseudo-rehearsal-based CL methods. GANs~\citep{goodfellow2014generative} and auto-encoder networks have been extensively used in this context. While pseudo-rehearsal methods offer an effective means to reduce storage requirements, the generative model employed in these methods requires significant storage for its parameters. Challenges in convergence and the potential occurrence of mode collapse may make them less optimal for online CL. Several other contributions in this segment include~\cite{shin2017continual, kemker2017fearnet, seff2017continual, van2018generative, wu2018memory, van2020brain, pomponi2020pseudo, wang2021triple}.

EWC~\citep{kirkpatrick2017overcoming} and pseudo-rehearsal has been adapted to overcome CF in continual image generation scenarios in GANs by~\cite{seff2017continual}. Another similar work is MeRGANs~\citep{wu2018memory}, a coalition between a memory replay generator and a conditional GAN framework. The generator replays past memories (via generative sampling) to overcome CF while learning the latest task. They introduce two variations: (i) Joint training - replaying samples during training and (ii) Replay alignment - outputs from the previous generator and the current generator are made to synchronize.
The application of pseudo-rehearsal for image classification was tested by~\cite{shin2017continual} (DGR), where a classifier followed an unconditional GAN. Another closely related work is FearNet~\citep{kemker2017fearnet}, a brain-inspired threefold, generative autoencoder-based class incremental learning mechanism. It functions using a short-term memory system, long-term memory storage system functioning with pseudo-rehearsal strategies to consolidate memory from previous tasks and an intermediate system that chooses from between the memory systems during inference.

% \clearpage
% \MH{my version below}
\begin{algorithm}
    \SetAlgoLined
    \footnotesize
    \caption{Pseudo-code for generative replay-based CL}\label{alg:gen-replay}
    
    \KwInput{New task training data $\mathcal{D}_i = \{(\vx_j^{(i)},\vy_j^{(i)})\}_{j=1}^{m_i}$,\\ 
        initial parameters $\vtheta \in \mathbb{R}^n$, \\
        initial parameters for the generative model $\phi \in \mathbb{R}^n$, \\
        replay buffer $\mathcal{B}$, batch size $b \in \mathbb{N}$, \\
        number of epochs $ep$.}        
    \KwOutput{Updated parameters $\vtheta^\ast \in \mathbb{R}^n$.}

    \For {each task $i$} {
        \For {each epoch $e = 1$ \KwTo $ep$} {
            \For {each mini-batch} {
                $\mathcal{D}_1 \gets \texttt{Sample}(\mathcal{D}_i, b)$\;
                $\mathcal{D}_2 \gets \texttt{Sample}(\mathcal{B}, b)$\;
                $\mathcal{D}_{batch} \gets \mathcal{D}_1 \cup \mathcal{D}_2$\;
                $\hat{\vy} \gets f(\vx; \vtheta)$ for $(\vx, \vy) \in \mathcal{D}_{batch}$\;
                $\mathcal{L}(\vtheta) \leftarrow \mathbb{E}_{(\vx,\vy) \sim \mathcal{D}_{batch}} \texttt{Loss}(\vy, \hat{\vy}; \vtheta)$\;
                $\vtheta \leftarrow \vtheta - \eta \nabla_\vtheta \mathcal{L}(\vtheta)$\;
            }
        }
        
        % Generate synthetic samples
        $\mathcal{B}_s = \{(\vx_s,\vy_s)\}_{s=1}^{m} \leftarrow f(\vx; \phi)$ ; Generate synthetic samples using the generative model\;
        
        % Update the buffer
        $\mathcal{B} \gets \mathcal{B} \cup \mathcal{B}_s$ ; Add synthetic samples to the buffer\;
    }
\end{algorithm}
% \clearpage

\textbf{Hybrid replay methods} utilize both rehearsal and pseudo-rehearsal for optimal performance. The main challenges to tackle are the additional memory that needs to be maintained with the increasing number of tasks and to address privacy concerns in certain usecases. Auto-Hetero (AH)~\citep{solinas2022beneficial} focused on enhancing classification accuracy through reinjections from random noise (iterative sampling) used in the hybrid pseudo-sample generation process. Their findings show that hyperparameters like the buffer size, strength of the noise, and the number of reinjections affect the successful prior knowledge retrieval.

In conclusion, the CL landscape is multifaceted, with a range of methodologies, challenges, and numerous recent advancements. The literature on CL methodologies has expanded in recent years, introducing novel approaches. Several studies~\citep{mirzadeh2020understanding, mirzadeh2022wide, mirzadeh2022architecture} have been carried out to understand different factors affecting the CL nature of NNs. Although the growth in supervised CL context is remarkable, these methods are not directly applicable to CL in RL setups, which remains a nuanced challenge.

\subsection{Reinforcement Learning}
\label{sec:RL}

RL differs from conventional supervised learning as it relies on a trial-and-error method of learning, making it suitable for a wide range of real-world applications where prior collected datasets aren't available in hand. In online RL, an agent is not informed about the optimal actions in its long-term interests after taking an action. Therefore, the agent must acquire valuable experience and learn the optimal and meaningful sequence of actions through interactions with and respective feedback from the environment.

%notation
\subsubsection{Markov Decision Process (MDP)}
\label{lab:mdp_pomdp}

\begin{figure*}[!b]
\centering
\footnotesize
\begin{tabular}{{c@{ } c@{ } c@{ }}}
     {\includegraphics[width=0.45\linewidth, valign=t]{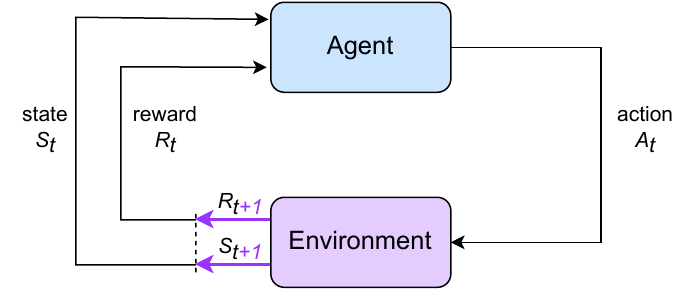}}&
     \hspace{0.15cm} &
    {\includegraphics[width=0.5\linewidth, valign=t]{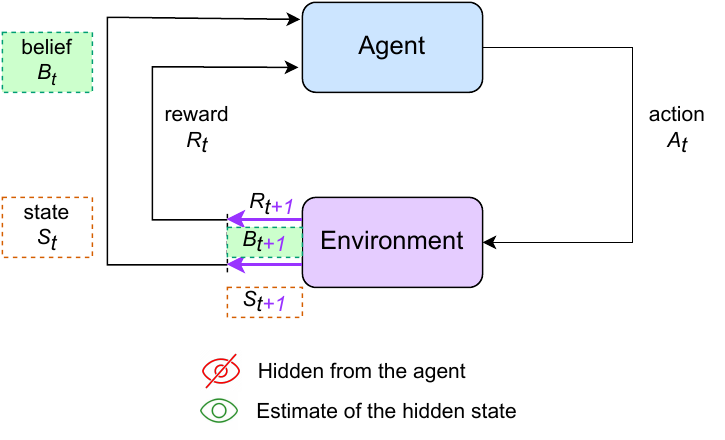}}\\
    (a) && (b)\\
  \end{tabular}
\caption{Interaction between an RL agent and MDP/POMDP based environment. (a) The interaction within an MDP~\cite{sutton2018reinforcement} (b) The interaction within a POMDP}
\label{fig:mdp_pomdp}
\end{figure*}

MDPs are mathematical models utilized in the decision-making process of a stochastic environment. These models help define the RL agents' interaction with the environment. An MDP is defined as a tuple \mbox{$\mdp = (\states,\actions,\transitions,\initstate,\rewards,\discount)$} where $\states$ is a set of states, either discrete or continuous, and $\actions$ is a set of actions, which similarly can be discrete or continuous.
% $\bs \in \states$
%  $\ba \in \actions$
Transitions $\transitions$ define a conditional probability distribution of the form $\transitions(\bs_{t+1} | \bs_t, \ba_t), \bs_{t+1}, \bs_t \in \states, \ba_t \in \actions$ that describe the dynamics of the system, and $\initstate$ defines the initial state distribution $\initstate(\bs_0)$. 
The reward function is defined by $\rewards: \states \times \actions \rightarrow \real$, while $\gamma \in (0, 1]$ being the discount factor quantifies the importance given for future rewards. 
The probabilistic transition from state $\bs_{t}$ to $\bs_{t+1}$ depends on the action taken from the set of available actions at state $\bs_{t}$. The agent's goal is to maximize the rewards it receives from the environment in the long run (by taking appropriate actions at each state). This is generally known as maximizing the expected return $\return_{t}$, 
\begin{equation}
    \return_t = \sum^{\infty}_{k=0} \gamma^k\rewards_{t+k+1}
\end{equation}
~\citep{puterman2014markov, sutton2018reinforcement, levine2020offline}. 

It is useful to note that the MDP itself defines a continual learning problem, the dynamics are stochastic and not necessarily stationary, as are the rewards. 

% States specify the current location of the agent in an environment, while actions are the movements and functionalities (moving sideways, vertically, jumping, shooting, etc.) an agent can perform. The probabilistic transition from state $\bs_t$ to $\bs_{t+1}$ depends on the action taken from the set of available actions at state $\bs$. This leads to receiving a reward which could fluctuate between positive, negative or zero values depending on the reward function definition of the environment~\cite{sutton2018reinforcement}.
The Markov property is the key assumption in MDPs. The assumption states that future states and rewards are only dependent on the current state and action, thus not dependent on past states or actions. This alleviates the need to maintain a history of past states and actions. Due to the sequential decision-making pattern, current actions influence not only immediate rewards but also affect future events or states and, through them, future rewards~\cite {sutton2018reinforcement}. The policy ($\policy: \states \times \actions \to [0,1]$) maps each state to probabilities of selecting each viable action the agent may take while visiting these particular states. The objective of an RL agent is to learn the optimal policy ($\policy^\ast$) that, in turn, maximizes the expected cumulative reward over time. The RL agent following a particular learning mechanism will assist in this process through trial and error learning.

RL-based algorithms in the field of machine learning tend to use the scalar reward feedback received to guide the agent to learn how to achieve defined goals in an environment~\citep{van2012reinforcement}. The RL system mainly comprises the agent, which has the decision-making and learning ability, and everything apart from the agent belongs to the environment; see \Cref{fig:mdp_pomdp} for a conceptual diagram. 

%MDP, POMDP
In fully-observable MDPs, the agent has full knowledge of its current state to make decisions to learn an optimal policy compared to POMDPs (Partially-Observable Markov Decision Processes) where the information provided by the environment is partially observable and hence insufficient to make a decision with high certainty. This results in a generalized MDP model allowing other forms of uncertainty to be considered in the decision-making procedure~\citep{cassandra1998survey}. This means the true state is unknown, but a belief of the true state using observations could be known. For instance, in robotics, the probability distribution functions over states, referred to as beliefs, define the robot and environment state and policies determine the optimal actions to take based on these beliefs rather than deterministic states~\citep{kurniawati2022partially}.

A POMDP is a more general framework defined by {$(\states,\actions,\transitions,\rewards,\observations,\observationFunction)$} where the only addition to the MDP definition are $\observations$, a finite set of observations and $\observationFunction: \states \times \actions \times \observations \rightarrow [0,1]$ observation function, capturing errors and noise in measurement and perception. $\observations(\bs_{t+1},\ba_{t},\bz_{t}) = P(\bz_{t}|\ba_{t},\bs_{t+1})$ specifies the likelihood of observing $\bz$ when action $\ba$ is executed, leading to state $\bs_{t+1}$~\citep{ng2010towards}. Solving a POMDP-based problem involves determining an optimal policy ($\policy^*$), which is a mapping from beliefs to actions $\policy^*: \belief \rightarrow \actions$ aimed at maximizing the objective function~\citep{kaelbling1998planning}.

\subsubsection{On-policy RL vs Off-policy RL vs Offline RL}
\label{sec:on_off_offline}
RL agents focus on learning actions that optimize their behavior in the environment while trying to act randomly or following a targeted acquisition approach to explore different options. Learning can be performed in multiple ways, determined and conditioned primarily on how experiences are generated.

In \textbf{on-policy} learning, the most recently acquired policy guides the repetitive process of collecting experience, evaluating, and self-improvement. This involves applying actions based on a nearly optimal behavior policy that emphasizes exploration~\citep{sutton2018reinforcement}. Alternatively, \textbf{off-policy} learning is an approach that involves accumulating, storing, and updating experiences in a replay buffer while the agent interacts with the environment. Training the policy and updating it is achieved by sampling from this resource. In \textbf{offline} RL, the agent is barred from real-time interactions for data gathering. Instead, the focus is on leveraging pre-collected data samples to learn the best possible policy. \Cref{tab:on_off_offline_rl} outlines the key considerations to take into account when selecting a learning mechanism for a specific RL problem.

\begin{table}[h!]
    \centering
    \footnotesize
    {\renewcommand{\arraystretch}{2}%
    \begin{tabular}{|p{2.5cm}|p{3.6cm}|p{3.6cm}|p{3.6cm}|}
    % \begin{tabular}{| >{\raggedright}p{3cm}| >{\raggedright}p{4.2cm}| >{\raggedright}p{4.2cm}| >{\raggedright}p{4.2cm}|}
    % \begin{tabularx}{\textwidth}{L{0.4} L{0.7} *{3} }
        \hline
         \Centering{\textbf{Key factors}}& \Centering{\textbf{On-policy RL}} & \Centering{\textbf{Off-policy RL}} & \Centering{\textbf{Offline-RL}}\\
         \hline
         \RaggedRight{Exploration vs Exploitation} &\RaggedRight{Essential since actions influence the data distribution used to learn the policy}  &\RaggedRight{Favored for achieving a balance between exploration and exploitation as it efficiently leverages past experiences.} &\RaggedRight{Highly appropriate when exploration is restricted} \\
         \hline
         Data efficiency &\RaggedRight{Data inefficient since relies on latest rollout for learning and discards collected experiences} &\RaggedRight{Moderately data efficient since reuses past experiences}&\RaggedRight{No new data collection but challenged if the dataset does not cover the entire state-action space adequately} \\
         \hline
         \RaggedRight{Stability and convergence}& \RaggedRight{More stable due to the consistency between the data collection policy and the policy being optimized} & \RaggedRight{Potential instability, as it involves optimizing a policy different from the one used during data collection} & \RaggedRight{Stability, depends on the quality and representativeness of the pre-collected data} \\
         \hline
         \RaggedRight{Real-time constraints}&\RaggedRight{Suitable for online interaction} & \RaggedRight{When real-time interactions are limited} & \RaggedRight{When real-time interactions are unavailable}\\
         \hline
         \RaggedRight{Resource constraints}& \RaggedRight{Resource intensive due to continuous exploration} & \RaggedRight{Resource efficient due to data reuse} & \RaggedRight{Resource efficient if data collection and learning are decoupled} \\
         \hline
         Examples & \RaggedRight{Robot arm control \citep{schulman2015trust, schulman2017proximal}, Massively parallel RL~\citep{rudin2022learning}} & \RaggedRight{Autonomous driving \citep{hasselt2010double, mnih2013playing}} & \RaggedRight{Robotics and medical diagnosis where online interactions are expensive or impossible~\citep{levine2020offline}}\\
         \hline
         RL Algorithms & \RaggedRight{PPO~\citep{schulman2017proximal},A2C/A3C~\citep{mnih2016asynchronous}, TRPO~\citep{schulman2015trust}} & \RaggedRight{DQN~\citep{mnih2013playing}, HER~\citep{andrychowicz2017hindsight}, SAC~\citep{haarnoja2018soft}, TD3~\citep{fujimoto2018addressing}} & \RaggedRight{MBMF~\citep{nagabandi2018neural}, I2A~\citep{weber2017imagination}}\\
         \hline
    % \end{tabularx}
    \end{tabular}}
    \caption{Comparison of On-Policy, Off-Policy, and Offline RL methods based on key considerations}
    \label{tab:on_off_offline_rl}
\end{table}

Choosing a learning methodology from within on-policy, off-policy, or offline RL in a CL setup depends on multiple factors, such as the nature of the learning environment, the availability of resources, and the characteristics of the specific task. On-policy methods can successfully cater adaptation to non-stationary environments but could perform inefficiently due to additional exploration. Under multi-task training scenarios, implicitly parallelising policy learning across simulated environments has shown great potential to exploit computational resources~\citep{rudin2022learning}. Off-policy learning methods can learn efficiently in stable environments, although learning from dynamic setups could be challenging due to the reuse of past data samples~\citep{steinparz2022reactive}. Offline learning methods are advantageous when data collection through online interactions is expensive or risky (\eg robots, healthcare) or inefficient. They perform similarly to off-policy methods with the added disadvantage of the inability to further explore to gather new data.

In a CL setup, hybrid approaches combining multiple policy learning methods have also proven to be beneficial~\citep{rolnick2019experience}. For example, initially learning using on-policy methods to adapt quickly to new tasks and then using off-policy methods to efficiently leverage past experiences for more stable tasks could provide a balanced approach. 
% \AZ{should we add a small intro to Decision Transformers?}

\subsubsection{Model-free vs Model-based Learning}
In scenarios where individuals encounter novel and unfamiliar environments, the process of exploration becomes instrumental in acquiring insights and adapting to the surroundings. This adaptive behavior is particularly crucial when navigating uncharted territories, as it facilitates the accumulation of experiential knowledge and the discernment of appropriate actions. Let us consider an agent trying to navigate a maze. 
Under \textbf{model-free learning}, upon entering this novel environment, the agent has to explore and learn to make decisions based on the experience and feedback received by interacting with this environment. This includes using historical data of states, actions, and rewards with no explicit understanding of the underlying dynamics of the environment. Towards the end of the learning process, the agent learns the optimal policy to make favorable decisions, although it may not have a clear understanding of the maze's structure. 
Conversely, under \textbf{model-based learning}, the agent seeks to learn a model of the environment dynamics in an off-policy manner, helping it to predict future state changes in response to the agent's actions, thereby training a policy that optimizes the agent's behavior. Thus, they \textit{plan} the actions to take by simulating scenarios through the learnt model. After considering available options, the agent takes the action that leads to the best outcome~\citep{sutton2018reinforcement}.

Although model-free RL algorithms may face challenges with slow convergence and high variance due to the value estimations performed, they excel in settings challenging to craft accurately using a model or in situations where the dynamics remain uncertain. This is accomplished at the expense of being highly data-demanding, as the agent requires a multitude of interactions to learn an optimal policy~\citep{mnih2013playing}. This limits the application of model-free RL methods to mechanical (robotic) systems that may wear-out or require regular environment resetting, in comparison to video games or simulated environments~\citep{deisenroth2011pilco}. Model-based RL algorithms, on the other hand, learns the dynamics of the environment efficiently and accurately, while being highly sample efficient since if necessary, additional data could be generated through the learnt model of the environment~\citep{clavera2018model, ebert2018visual}. However, in this setting, the quality of the policy learnt depends on the accuracy of the learned model.

In the context of CRL, the expectation is to learn seamlessly across non-stationary data distributions that represent evolving environments, shifting tasks, or changing objectives, enabling the model to adapt continuously without forgetting previously acquired knowledge. A CRL agent should be capable of adapting to these changes regardless of having an exceptional knowledge of these environments. This indeed represents the practical challenges in the world, say, for a robot. Consequently, balancing the trade-offs between sample efficiency, adaptability, and robustness is crucial. Whether employing model-based or model-free methods, the ultimate goal remains the same: to develop agents that can learn and adapt continually, mirroring the dynamic and unpredictable nature of real-world environments.

\subsubsection{Learning Environments}
\label{sec:rl_envs}
RL involves training an agent that learns to achieve goals using the feedback received through continuous interactions with an environment. In order to compare the performance of agents following different learning algorithms, it is vital to test them with appropriate test beds.

Existing environments have been instrumental in the development of single-task RL algorithms~\citep{berner2019dota, vinyals2019grandmaster} learning  stationary policies. Physical simulators combined with gaming environments provide partial observations and multi-observation modes favorable for CL scenarios, although they lack complexity~\citep{johnson2022l2explorer}. A significant challenge in conducting studies for CRL is the absence of standard, widely accepted, and configurable environments to evaluate the performance of agents through carefully designed experiments. Many recent works often prioritize the careful selection of environments/tasks to address specific problems. These practices often do not reflect the practical complexity and unpredictability encountered in real-world situations. Although in CRL, Atari~\citep{bellemare2013arcade} has been the widely used environment, lately it has been less preferred specifically due to its sample inefficiency and lack of varying levels of difficulties~\citep{wolczyk2021continual}.
Many new test beds have emerged recently to tackle various shortcomings of existing resources. The following section will discuss these in detail with respect to their characteristics.

\begin{figure}[h]
    \centering
    \includegraphics[width=1\linewidth]{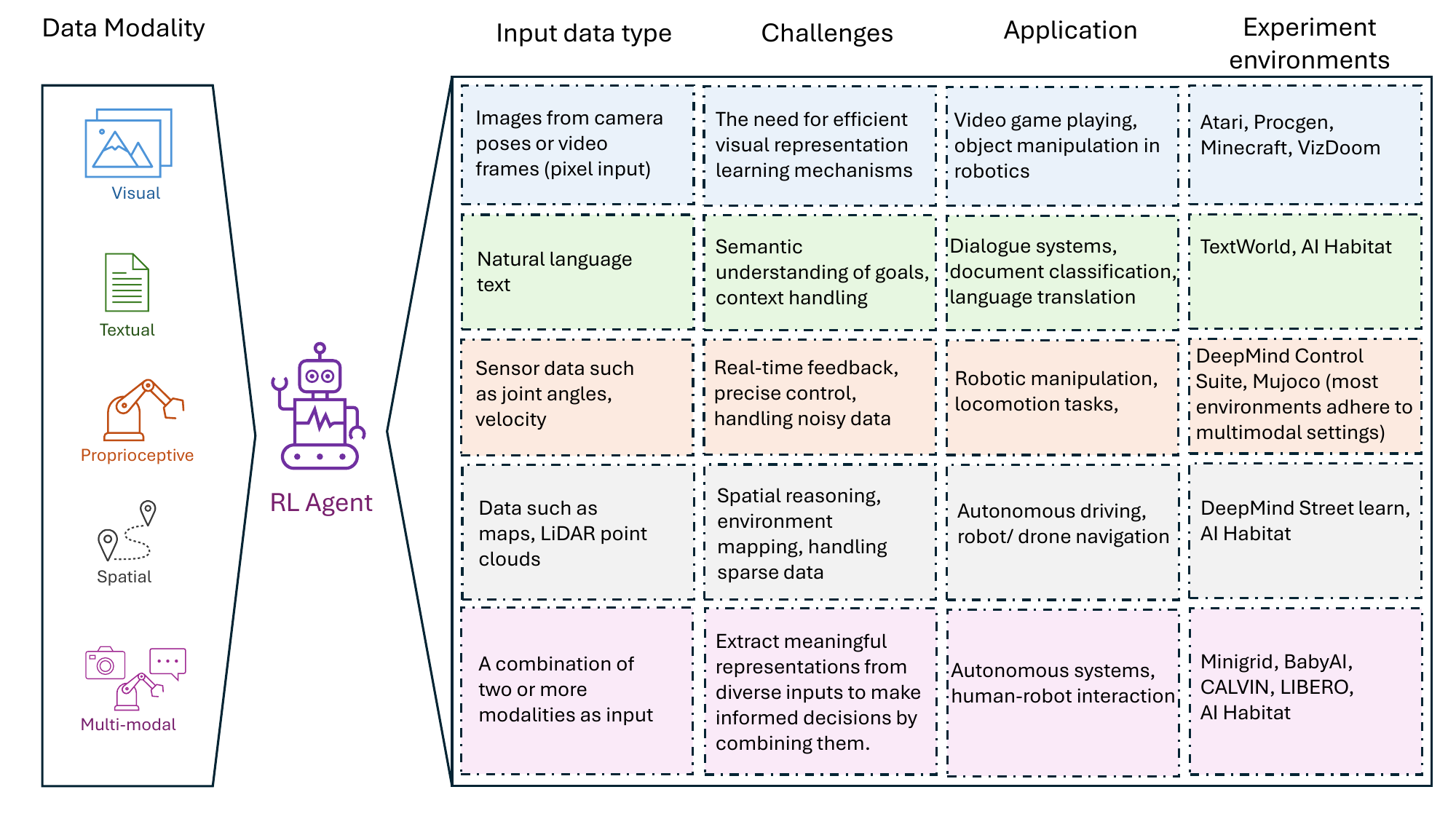}
    \caption{A summary on data modality-based inputs for an RL agent.}
    \label{fig:rl-modality}
\end{figure}

Some environments have predefined tasks that agents can perform without altering configurations. In contrast, others permit changes to visual features, such as colors and difficulty levels, challenging the learning agent and providing opportunities to evaluate its performance under various aspects. DeepMind Lab~\citep{beattie2016deepmind} is a 3D environment capable of generating realistic visuals in 3D space, while Procgen~\citep{cobbe2019procgen} and Minigrid~\citep{MinigridMiniworld23} consists of environments defined in a 2D space. 
% The accessible actions in \Cref{fig:2d3d}(a) depict only forward and backward movement, whereas the additional jump and crouch options in \Cref{fig:2d3d}(b) indicate the existence of a 3D space. 
More examples and respective characteristics of learning environments used for CRL research have been summarised in \Cref{tab:rlenv-papers}. A detailed and organized list of existing RL environments has been discussed in \href{https://github.com/clvrai/awesome-rl-envs}{this} GitHub repository from which the reader could choose options for single task RL or CRL contexts. Below in \Cref{fig:rl-modality} we summarize commonly used RL environments based on the modality of the input features provided to the agent. As depicted, selecting the appropriate RL algorithm to address diverse challenges in these environments is complex and requires careful consideration of various factors, including the nature of the input modalities and the specific characteristics of each environment.

% \paragraph{Characterisation of RL Environments}
% The complexity and visual appearance of the learning environments directly affect the policies learnt by the RL agents and thereby their final performance. It is vital to be aware of the dynamics of such learning environments before applying them to CL scenarios. 

\paragraph{Offline-RL Environment Datasets}
As detailed in \Cref{sec:on_off_offline}, when crafting solutions with offline RL, the established practice for obtaining a comprehensive dataset involves extracting trajectories from an expert agent trained on the environment or using a replay buffer from an online RL algorithm (such as in DAgger~\citep{ross2011reduction},~\cite{fujimoto2019off, kumar2019stabilizing, agarwal2020optimistic}). It has been recognized that certain existing benchmarks are not specifically tailored for offline settings, leading to challenges in conducting accurate performance evaluations~\citep{gulcehre2006rl, schrittwieser2021online}. Additionally, some agents claiming expertise may be only partially trained, lacking representations that encompass the entire state-space of the given environment. To address these challenges,~\cite{fu2020d4rl} has released an open-source benchmark that caters to the essential characteristics required for offline RL in real-world applications, along with corresponding methods for evaluating their performance. Another offline RL dataset based on the continuous control benchmark problem was made public by ~\cite{mitchell2021offline} where existing variations include varying goal directions and velocities.

\subsubsection{Algorithmic Approaches in Continual Reinforcement Learning}

In CRL, algorithms from all variations of RL (see \ref{sec:on_off_offline}) have been applied along with CL methodologies to enable AI agents to learn continually in non-stationary environments. 

Under model-free RL, Q-learning~\citep{watkins1992q}, is an off-policy RL method that learns the optimal action-value function $Q_\policy(\bs,\ba)$ (Q-function) that estimates the expected total reward of taking a particular action in a specific state while following the optimal policy thereafter. The algorithm tries to trade off between exploiting the current knowledge and exploring to gain new knowledge. The $Q_\policy(\bs,\ba)$ update is based on the Bellman equation~\citep{bellman1966dynamic}. The Q-table (estimating Q-values) was replaced by a function approximator when transitioning from Q-learning to its parametric implementation using NNs, Deep Q-Networks~\citep{mnih2013playing}.

% \AZ{Add \cite{mankowitz2018unicorn}}

Examples of the Q-Learning based RL algorithms tested for CL include;
\begin{itemize}
    \item DQN~\citep{mnih2013playing} - Learns utilizing samples/experiences saved in a replay buffer and a target Q-network.
    \item Rainbow~\citep{hessel2018rainbow} - An integrated improvement over the off-policy DQN algorithm.[ref~\citep{agarwal2022reincarnating}]
\end{itemize}
    
Policy gradient-based methods select actions by learning a parameterized policy. Here, the goal is to control the probability distribution that represents the actions for the given state in a manner such that positive actions (that reap more rewards) are sampled more often in the future. The algorithms under this section can belong to on-policy or off-policy methods, which differ by the necessity of providing experience obtained by following the current policy or following from a different policy than the one being followed. Hence, data efficiency plays a major role since samples are discarded after each model update step when using on-policy algorithms.

Some of the frequently used policy-gradient algorithms under this sector include;
\begin{itemize}
    \item PPO~\citep{schulman2017proximal} - (On-policy) Controls the policy update by maximizing a clipped surrogate objective function. Unlike TRPO~\cite{schulman2015trust}, which explicitly constraints the policy update by enforcing a limit on the KL divergence - a measure of how much the new policy diverges from the old one.
    \item A3C~\citep{mnih2016asynchronous} - (On-policy) Consists of an Actor (Policy network) that upgrades the policy gradients to improve decisions taken on the actions for each state while the Critic (Value network) criticizes actions taken by the Actor based on a state-value calculation.
    \item SAC~\citep{haarnoja2018soft} - (Off-policy) Learns in continuous action spaces aiming to maximize the expected cumulative reward. Exploration is encouraged through a maximum entropy regularisation term while it learns with the help of a replay buffer~\citep{xie2022lifelong, wolczyk2022disentangling}.
\end{itemize}

Despite the impressive and successful achievements deep RL agents have demonstrated over the years~\citep{mnih2013playing, moravvcik2017deepstack, silver2017mastering, finn2017model}, there have been issues stemming from network design choices~\citep{henderson2018deep}, reproducibility concerns~\citep{colas2018many}, and the reliability of results~\citep{rl_reliability_metrics} \etc. More recent studies carried out on deep RL algorithms have raised concerns that the the need to constantly improve it's policy and/or value functions to match non-stationary target functions affects the brittle nature of these algorithms~\citep{lyle2022understanding}. These challenges are compounded by issues related to loss of capacity~\citep{lyle2022understanding}, loss of plasticity~\citep{lyle2023understanding, abbas2023loss, nikishin2024deep} and primacy bias~\citep{nikishin2022primacy} hindering the ability to scale and adapt to the to the complex and dynamic nature of real-world environments.

\section{Continual Reinforcement Learning}
\label{sec:CRL}

% The definition of CRL problems initiate with a set of $\numtasks$ tasks of the sequence $\task_{1}, \task_{2}, ..., \task_{\numtasks}$ where each task $\task_{i}$ follows a POMDP framework~\cite{caccia2023task} {$(\states_{i},\actions_{i},\transitions_{i},\reward_{i},\observations_{i},\observationFunction_{i})$}
% % \mbox{$\mdp_{i} = (\states_{i},\actions_{i},\transitions_{i},\reward_{i},\discount)$} 
% (check~\cref{sec:RL} for notations). Instances where the entire state space is visible to the agent defines an MDP-based CRL problem, a special case of POMDPs. The goal in CRL is to learn a global policy $\policy$, that adapts to the continually changing POMDP-based tasks while mitigating CF being given or denied access to the task identifier $i$.

Continual Reinforcement Learning entails an RL agent learning to acquire an adaptive policy $\policy$ for optimal performance across a series of sequentially presented tasks $\task_{1}, \task_{2}, \ldots, \task_{\numtasks}$ experienced one at a time (where total number of tasks is $\numtasks$). The agent faces the challenge of mitigating catastrophic forgetting of previously encountered tasks caused by weight updates during backpropagation in neural networks. Each of these tasks is presumed to follow the framework of a POMDP~\citep{caccia2023task} {$(\states_{i},\actions_{i},\transitions_{i},\rewards_{i},\observations_{i},\observationFunction_{i})$} (see \Cref{lab:mdp_pomdp}), which provides a structure for addressing non-stationarity and partial observability in sequential decision-making problems.  Instances where the entire state space is visible to the agent defines an MDP-based CRL problem, a special case of POMDPs.

Some of the common assumptions held in CRL are that:
\begin{itemize}
    \item The data samples arriving are not i.i.d.
    \item Data from the previous task is not available at later tasks
    \item The tasks sequentially exposed to the agent are formulated as a POMDP
    \item The agent has no authority to choose the next task inline
    \item The risk of catastrophic forgetting exists
    \item Knowledge transfer is possible
\end{itemize}
% It is also expected that the range of experiences the RL agent requires, to perform optimally in an environment to be improved by the experiences collected from earlier tasks. 
It is also expected that leveraging the knowledge and experiences accumulated from earlier tasks would enrich the performance on the current task. To achieve state-of-the-art performance in this scenario, CL is the ideal method to learn sequentially with minimal performance loss on past tasks.

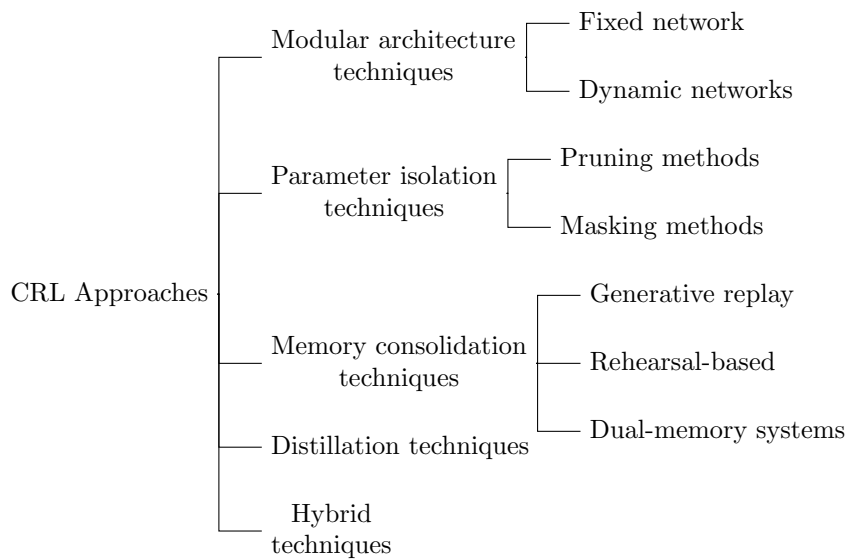
\begin{figure}[h!]
    \centering
    \footnotesize
    \small
    \begin{forest}
        for tree={
            grow'=0,
            parent anchor=east,
            child anchor=west,
            anchor=west,
            edge path={\noexpand\path[\forestoption{edge}] (\forestOve{\forestove{@parent}}{name}.parent anchor) -- +(0.5pt,0) |- (\forestove{name}.child anchor)\forestoption{edge label};},
            align=center,
            l sep+=5pt,
        }
        [CRL Approaches
            [Modular architecture\\ techniques
                [Fixed network]
                [Dynamic networks]
            ] 
            [Parameter isolation\\ techniques
                [Pruning methods]
                [Masking methods]
            ]
            [Memory consolidation\\ techniques
                [Generative replay
                ]
                [Rehearsal-based]'
                [Dual-memory systems]
            ]
            [Distillation techniques]
            [Hybrid\\ techniques]
        ]
    \end{forest}
    \caption{A vertical tree-based representation of the various families or methods in CRL and the distinct branches within each family.}
    \label{fig:CRL_breakdown}
\end{figure}

In this section, we aim to explore and discuss some of the seminal CRL works that have paved the way for various strategies to address CF. These works have introduced innovative approaches to foster both forward and backward transfer of knowledge, thereby contributing significantly to the advancement of continual learning paradigms. The literature on CRL can be branched as shown in \Cref{fig:CRL_breakdown}. 

% \cite{luketina2022meta}

\subsection{Task-aware vs Task-agnostic Learning}
\label{sec:taw_tag}

A distinction should be made between settings where the transition between tasks is known (task-aware) and unknown (task-agnostic). This may depend on the problem setting.
%Acknowledging the transition between tasks in the CL setting is not universally optional. Its necessity depends on the specific problem at hand; while some problems inherently provide this awareness, others do not.\AZ{how about now?}

% \sout{Providing awareness to the agent on the transition between tasks in the CL setting is optional.}
% \MH{not optional, it is problem-dependent, for some problems you have that knowledge, for some you don't}

Instances where the transition is identified using an external change in Task index (Task-ID) or identified by external means fall under the \textbf{task-aware} CRL setting. In these instances, learnings from the past are consolidated with the knowledge acquired in the latest task as the agent moves into the next task (is informed about the end of the current task). In \textbf{task-agnostic} learning, the agent is responsible for detecting the transition from one task to another without external instructions or interventions to better coordinate its knowledge reserve. To do so, it should be able to identify significant shifts in either the reward function, observations, set of actions to take, or any other combination of environmental dynamics~\citep{steinparz2022reactive}. The progress made in task-agnostic learning has been limited, primarily due to the inherent challenge of recognising task transitions, especially in the context of RL. Since RL agents are designed to adapt to new environments, it is vital in CRL to recognise the moment the environments/tasks transition and accordingly consolidate knowledge relevant to past tasks. 
While the primary objective is to solve multiple tasks, task-aware methods strive to memorize the optimal policy for each task through parameter sharing. Conversely, task-agnostic methods aim to acquire a universal solution that can swiftly adapt to both new and existing tasks, incorporating a task inference technique~\citep{caccia2023task}.

A recent study by~\cite{caccia2023task} explored the factors contributing to the performance variation between task-agnostic CL and multi-task agents in an RL setting. They observe two advantages of using off-policy RL methods in task-agnostic settings. Sample efficiency is preferred in CRL since agents spend limited time in each environment and rarely revisit these. Next, the decoupling of the actor learning from the policy provides the opportunity for replay. The recurrent memory adjoined task-agnostic continual learner (Replay-based Recurrent RL (3RL)) introduced in this study confirmed their hypotheses that \textbf{(i)} Task agnostic techniques are advantageous when data and computation are limited or data is of high dimensionality \textbf{(ii)} Fast adaptation in task-agnostic settings mitigates CF. An observed performance increase where a CL agent surpassed an equivalent multi-task learner was credited to reduced gradient conflict in 3RL.

\subsection*{Change-point detection}
\begin{figure}[h]
    \centering
    \includegraphics[width=1\linewidth]{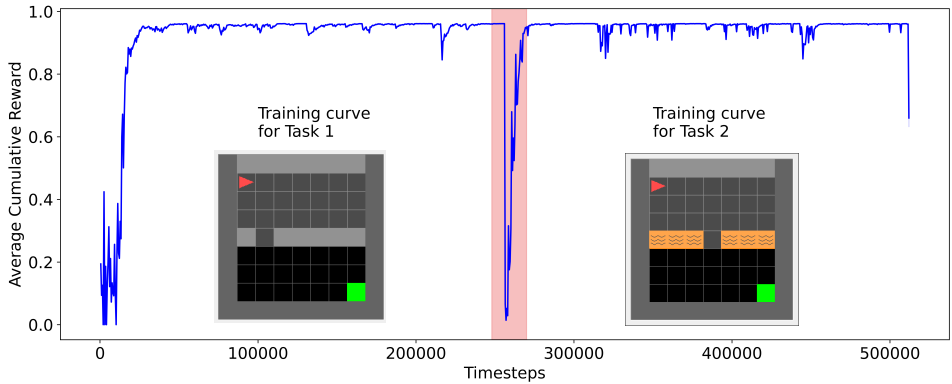}
    \caption{An illustrative representation showing a change-point (highlighted in red) resulting from a domain shift while adapting to sequentially navigate between two Minigrid~\citep{MinigridMiniworld23} environments, where observations undergo alterations.}
    \label{fig:changepoint}
\end{figure}

% \newline
Change-point Detection is one approach to cope with non-stationarity in a lifelong RL context where the domain shift specifically occurs in reward functions or transition dynamics. Since no prior warning is issued to the learning agent, it is vital to make provisions to detect such shifts and adapt the policies to new domains. When RL agents are equipped with this ability, task-agnostic CL can efficiently navigate sequential tasks with varying dynamics and reward structures, dynamically adjusting policies in response to unforeseen changes. This adaptability enhances the agent's performance in lifelong reinforcement learning, allowing it to effectively tackle non-stationarity and maintain effective decision-making across diverse and evolving environments. A visual example is provided in \Cref{fig:changepoint}. 

According to~\cite{steinparz2022reactive}, Reactive Exploration enables an agent to be motivated to re-explore segments of the environment when it recognises the change in one or more of the RL environment dynamics. The exploration mechanism is based on the curiosity model~\citep{pathak2017curiosity} where a forward and an inverse dynamics model identifies non-stationarities in the observations. The rewards model introduced in the paper handles the reward function changes. Although CL is not inherently present in the on-policy, PPO~\citep{schulman2017proximal} and off-policy, DQN~\citep{mnih2013playing} agents tested in this case, the proposed exploration mechanism addresses task agnostic CRL approaches for the on-policy case successfully.

Task agnostic approaches in rehearsal-based learning methods (see~\Cref{sec:memory_consolidation}) have also introduced domain shift recognition modules to adapt the generative models to new tasks. For example, a Welch's \textit{t}-test~\citep{welch1947generalization}, assesses the null hypothesis $H_{0}$  whether two populations have similar means. It assumes independence in the observations, non-significant outliers, non-identical population variances and normality in both the data distributions. This statistical test has been preferred in~\cite{caselles2018continual} due to its invariant nature to various scales when applied to identify the shift in the reconstruction error distribution of Variational Auto Encoders (VAE).  

\subsection{A Breakdown of Knowledge Retention Methods in CRL}

\subsubsection{Memory Consolidation Techniques}
\label{sec:memory_consolidation}

Memory consolidation is the transfer of memory from the hippocampus (short-term memory) to the cortex (long-term memory)~\citep{tse2007schemas, squire2015memory}. It has been found that while resting/sleeping, the hippocampus tends to replay patterns of activation similar to what occurred during the awake time to interleave prior experience into a more stable and comprehensive internal representation alleviating forgetting~\citep{louie2001temporally, wang2019boosting, hayes2021replay}. Algorithmic approaches under this segment try to maintain separate modules to acquire new knowledge and later transfer it to a permanent memory base, replicating the mammalian brain or continuing to replay past samples to avoid forgetting them while learning to solve new tasks. 

Progress and Compress (P\&C)~\citep{schwarz2018progress} is a learning system based on knowledge distillation. It effectively overcomes forgetting of previous tasks by the use of a knowledge base, controlled using a modified EWC~\citep{kirkpatrick2017overcoming} (see~\ref{para_regulaz_CL}) implementation (named Online EWC) during the \textit{Compress} phase, while an active column {(implemented as a NN)} is focused on accumulating knowledge from the current task during the \textit{Progress} phase. Alternatively, training these independent networks enables forward transfer when encountering a new task.

The work by~\cite{caselles2018continual} is one of the earliest task-agnostic CRL implementations developed based on pseudo-rehearsal. A VAE-based State Representation Learning (SRL)~\citep{lesort2018state} model circumvents the shortcoming of forgetting via generative replay, while encoding efficient state representations to discover the optimal RL policy. Non-stationarity is introduced through changes in states while the action space is kept fixed. Welch's \textit{t}-test~\citep{welch1947generalization} applied on the reconstruction error distribution of the VAE helps in the change-point detection of environments, in this case, the color change of objects in a 2D grid world. The evaluation looked at the successful reconstruction of realistic states by the VAE and catastrophic forgetting. 

A model-free generative replay approach with observation and action replay overcoming forgetting and brain-inspired memory consolidation during wake up and sleep cycle has been proposed by~\cite{daniels2022model}. CRL using world models to generate past experiences for rehearsal-based learning has been carried out by~\cite{ketz2019continual} in a task-agnostic manner.

A similar work is Reinforcement-Pseudo-Rehearsal (RePR\textsuperscript{2})~\citep{atkinson2021pseudo}, a dual memory system that, instead of reserving raw samples for replay-based learning, generates samples representative of previous tasks via a generative model. The novelty here is that an RL algorithm handles short-term memory while a long-term memory model is trained via supervised learning using the short-term network. This overcomes interference between old and new tasks. Next, the exchange of knowledge between the memory modules is executed using actual samples rather than generated ones. On a positive note, there is no memory capacity overhead since raw samples are not stored.

Other replay-centric mechanisms in place to overcome forgetting of past tasks include~\cite{riemer2018learning, rolnick2019experience, berseth2021comps} where experience replay was employed for multi-task and meta-experience learning scenarios. MTR~\citep{kaplanis2020continual} presented a task-agnostic memory approach including a shared episodic memory model~\citep{sorokin2019continual} eventually tested on multi-task and continual settings. Since changes to the learning environment may arrive uninformed, to overcome the constraint of dealing with only two buffers, they employ multiple sub-buffers to accumulate experience from varying timescales. To maintain memory to an extended period of time, the power law from psychological studies has been adapted. Invariant Risk Minimization loss encourages the agent to learn invariant mappings across all environments to mitigate challenges stemming from out-of-distribution generalization.

An elegant alternative solution to circumvent the need to store or generate samples was introduced by~\cite{kaplanis2018continual} following inspiration from the biological synapse model, which captures synaptic plasticity over a timescale. By modeling each parameter of the RL algorithm as a dynamical system of interacting variables rather than a scalar value, the synapse model helps overcome CF and mitigate the abrupt erasure of prior knowledge acquired. It was found that the synapse model could retrieve past learnings faster than the control agent in a CL setup within two tasks and a single task. A task-agnostic policy consolidation mechanism was thus developed~\citep{kaplanis2019policy} inspired by~\cite{kaplanis2018continual} and~\cite{rusu2015policy} where the agent's policy was continuously consolidated over multiple timescales into a cascade of evolving hidden networks. The knowledge is distilled back into the policy network to retain its policy closer to where it was previously.

The development of AFEC (Active Forgetting with Synaptic Expansion-Convergence) \citep{wang2021afec} is inspired by active forgetting mechanisms in biological systems. This Bayesian continual learning approach prioritizes new task learning by giving low precedence to the posterior distribution containing knowledge of the old task, facilitated by a forgetting factor. By selectively forgetting information, AFEC serves as a plug-and-play method, showcasing performance improvements in supervised and RL contexts, particularly complementing regularization and replay-based techniques. There have been many other works on selective forgetting for CL but specifically tested on supervised learning setups such as~\cite{shibata2021learning, yin2021mitigating, liu2022continual}.
\newline

Learning to solve tasks using pre-collected datasets is standard in supervised learning, but in RL, this is known as \textit{learning in an offline manner}. Offline RL avoids direct interaction with the environment, relying on the provided datasets to develop policies for task-solving. In the context of CRL, learning from a sequence of offline datasets with limited memory presents a significant challenge.
\newline
~\cite{gai2023oer} discuss how experience replay, which is effective in online CRL, introduces distribution shifts in continual offline RL due to (i) the mismatch between the behavior policy and the learning policy and (ii) the selected replay buffer and its impact on the learned policy. To address the latter, they propose a model-based experience selection (MBES) method, which loads the replay buffer with the most relevant and informative episodes from the dataset. Additionally, they introduce a dual behavior cloning (DBC) architecture to mitigate optimization conflicts that arise when using behavior cloning in offline settings.
\newline
Building on this work,~\cite{huang2024solving} explore using a Decision Transformer (DT)~\citep{chen2021decision} as a continual learner in offline RL setups. While DTs, trained with a supervised learning backbone, are capable of ignoring distribution shifts, they are prone to CF when all network parameters are updated in a continual learning setup. To counter this, a Multi-head DT is proposed, storing task-specific knowledge to mitigate CF, along with distillation and selective replay mechanisms to enhance learning for the current task when sufficient memory is available. In scenarios where a buffer is unavailable, the low-rank adaptation (LoRA)~\citep{hu2021lora} DT technique is suggested, merging weights with minimal impact on performance while fine-tuning the MLP layer in DT blocks to adapt to the current task.

\subsubsection{Modular(Constructive/Destructive) Architecture Techniques}
A storage-intensive method used to retain experience from past tasks is by either saving independent models representing each task learned or by adding parts to a single NN such that the new layers or parameters only focus on the current task. The separation of structure encourages better stability and plasticity but requires engineering to forward-transfer knowledge between tasks. Relevant works include~\cite{rosenbaum2017routing, kirsch2018modular, cases2019recursive, yang2020multi, anand2024prediction}

Progressive Networks~\citep{rusu2016progressive} was a solution to challenges introduced by fine-tuning in CL, a widely accepted method to transfer learning between models/domains. Progressive Networks use a growing NN architecture to reduce intervention between parameters optimized for past tasks by freezing them while adding new parameters to acquire new knowledge. It uses lateral connections between layers to incorporate learnings from past tasks to current ones, promoting shorter learning times. This effectively promotes the forward transfer of knowledge.

An ensemble-based method in the context of task-agnostic learning has been tested by SANE~\citep{powers2022self}. The ensemble consists of multiple (NN) modules or RL policies that are activated only when the agent determines the close relevance of the current state and predicted rewards to existing similar policies. New modules are generated only if similar modules cannot be found in the existing ensemble. Otherwise, modules are combined to form a new module, and the less important one gets discarded. Hence, this dynamic ensemble method is highly appropriate for situations where task identity is not available during training and testing. The experiments on different levels of 3 Procgen~\citep{cobbe2019procgen} environments depict how it overcomes forgetting by only updating the relevant module with predefined and bounded resources. This modular ensemble algorithm is a hybrid version of a rehearsal-based method to overcome forgetting and regularization enforced using behavioral cloning and knowledge distillation to preserve past knowledge.

Apart from this, modularity could also be induced through composition.~\cite{mendez2022modular} discusses how compositional learning has parted ways with lifelong learning, although the same CL problem could be solved by combining and reusing knowledge stored in multiple modules or components. This includes learning skills as separate policies or components and later combining them to solve specific tasks~\citep{lee2018composing} or transferring prior learnt experience to efficiently learn downstream tasks~\citep{pertsch2021accelerating}. This has been more prevalent in multi-task robotic domains compared to CRL in non-robotic environments. LiSP, a non-episodic skill learning-based CRL approach, has been presented by~\cite{lu2020reset} that learns from both online and offline interactions. This setup has adapted a reset-free scenario to test the RL agents' ability to navigate and function in a long horizon of tasks using a model-based planning module. 
% Other works in composition lifelong RL includes 

Hierarchical RL (HRL) is another avenue to handle complex problems where temporally abstract actions are fused into standard RL procedures~\citep{barto2003recent, riemer2018learning}. Learning can take place at different levels of abstraction or hierarchy compared to the traditional state to action mapping taking place on a low level. In HRL, low-level policies are selected and executed based on the decisions of higher-level policies, enabling a notion of solving subtasks. A parallel concept also lies with human behavioral research as discussed by~\cite{shallice2011organisation}. A problem in multi-task learning that emphasizes skill discovery and subsequent reuse of these skills to solve downstream tasks has been presented by~\cite{ding2022hliferl}. The options framework~\citep{sutton1999between} has been used to extract skills from a pre-trained model, and subsequently, a master policy is focused on selecting skills/policies for the incoming tasks.

\subsubsection{Parameter Isolation Techniques}
Algorithms under this subcategory isolate the parameters of a fixed NN such that parameters are learned in a task-specific manner. This ensures stability in learned weights since, in most cases, optimal parameters for past tasks are frozen at the start of a new task, allowing separation and promoting plasticity. The challenge here lies in the forward and backward transfer of knowledge due to the restriction of gradient updates, 2 of the main expectations in CL desiderata. 

PathNet~\citep{fernando2017pathnet}, discussed under CL methodologies, has been successfully tested on supervised and RL settings that show positive transfer of knowledge for a sequence of four tasks. An alternative approach to parameter isolation involves employing a mask over a fixed base network in order to identify subnetworks~\citep{mallya2018piggyback, ramanujan2020s, wortsman2020supermasks, koster2022signing}. This method offers the advantage of avoiding direct updates to the weights of the fixed NN. Instead, the freedom to generate masks for each task is utilized, allowing for independent updates tailored to optimize learning for each task without interference. All of the aforementioned works have been instrumental in testing this idea only in a supervised learning context subjected to limitations in the forward transfer of knowledge.
% This idea was first introduced by SupSup~\cite{wortsman2020supermasks} for a supervised learning scenario where supermasks were learnt on top of a randomly initialized base network that ensured no forgetting and was capable of handling task aware and task agnostic setups. 
~\cite{ben2022lifelong} has extended this idea to an RL setup where modulating binary masks have been tested on discrete and continuous state space RL environments, leveraging knowledge from past tasks to accelerate future learnings employing linear combinations of task-specific masks. 

Policy subspaces~\citep{gaya2021learning} (inspired by subspaces of models in supervised learning~\citep{wortsman2020supermasks}) is a solution for online policy adaptation to overcome limitations such as fine-tuning in single policy learning. Each point in the policy subspace is isolated using significant parameter values. By learning a variety of policies, a reserve of task specific policies is built to later choose from in the case where an unseen task is presented.
Parameter isolation has been extended to incorporate adaptive parameter growth based on the task sequence by building upon subspaces of policies by~\cite{gaya2022building}. The decision to grow or prune the existing subspace is decided based on the performance improvement or loss encountered by adding a new anchor(parameters) to the simplex, to handle a new task similar or drastically different to existing policies.

\subsubsection{Distillation-based Approaches}
Methods employing distillation in learning are centered on transferring knowledge from a proficient model, commonly referred to as the `teacher,' to a learner model. The learner model assimilates crucial information into a more streamlined format, ensuring sustained performance while reducing complexity and capacity. Notably, distillation-based learning diverges from modular architecture approaches, primarily due to its distinctiveness and its lack of emphasis on preserving information in long-term memory. Imitation learning is also a closely related paradigm, although a model-based algorithm often acts as a teacher. Distillation has been rarely tested in RL compared to its applications in supervised learning context~\citep{buciluǎ2006model, hinton2015distilling, buzzega2020dark}. Among the initial studies on distillation for single-game and multi-game policy in RL was introduced by~\cite{rusu2015policy}. 
% This has inspired many works including \AZ{complete}

\subsubsection{Hybrid Techniques}
OWL~\citep{kessler2022same} addressed an important CL problem involving interference caused due to utilizing the same RL environment with different goals as different tasks. The off-policy RL algorithm incorporates EWC to prevent forgetting, with separate heads ensuring plasticity for multiple tasks. For more stability, the replay buffer is reset at the end of each task. To achieve task-agnostic behavior during test time, a multi-armed bandit (MAB) selects the most relevant policy based on TD error. It's important to note that MAB's policy selection may be influenced by task diverseness, which may affect the TD error.

Hypernetworks~\citep{ha2016hypernetworks} have been one of the efficient methods to generate weights for a target network/s trained with backpropagation. This enables weight sharing across layers of a target network. We refer the reader to the comprehensive review by~\cite{chauhan2023brief}, which has covered a wide spectrum on this topic. One of the earliest applications of Hypernetwork for CL using regularisation was implemented by~\cite{von2019continual} that generated task-specific weights controlled by the task identity, introducing complete parameter isolation through learnable task-specific embeddings in a supervised learning setting. To overcome forgetting of previously learned weights, the amount of weight change for the current task is calculated and then combined with the task loss during each iteration as a two-step optimization technique.~\cite{schopf2022hypernetwork} had successfully extended this work to a CRL scenario where it was tested on an on-policy RL algorithm for a robotic arm environment. 

A comprehensive study has been carried out by~\cite{wolczyk2022disentangling} to disclose how components (the actor, critic, exploration, data) of a SAC~\citep{haarnoja2018soft} agent support the transfer of knowledge between tasks in Continual World~\citep{wolczyk2021continual}. Their findings state that; \textbf{(i)} Performance and forward transfer improve when a non-random exploration policy is combined with EWC; \textbf{(ii)} Behavioral cloning provides better performance and forward transfer than replay methods along with regularization applied to the actor in longer task sequences; \textbf{(iii)} Regularizing the critic deteriorates performance. The findings have successfully enabled the improvement of the default SAC algorithm, making it resistant to forgetting and increasing forward transfer in longer task sequences for multi-head and single-head setups.

%sun-burst diagram
% \begin{onion}*{0.3in}
%     \annulus{0}{0}{90}[1]
%     \annulus{0}{90}{180}[2]
%     \annulus{0}{180}{270}[3]
%     \annulus{0}{270}{360}[4]
%     \annulus*[yellow]{1}{0}{120}[Y is yellow]
%     \annulus*[red!20]{4}{0}{135}[This is a bunch of bla bla]
%     \annulus*[blue!30]{3}{60}{180}[This is some longer text]
%     \annulus[cyan!20]{2}{45}{130}
%     \annulus[white]{3}{225}{315}[Line 1\\Line 2]
%     \annulus*[purple!20]{2}{165}{325}[more\\stuff]
% \end{onion}

% ADD THE METRICS USED IN PARTICULAR CRL PAPEWRS

% \begin{onion}*{0.7in}
%     \annulus{0}{0}{180}[On-policy]
%     \annulus{0}{180}{360}[Off-policy]
%     \annulus[purple!10]{1}{0}{60}[Regularization\\-based]
%     \annulus[purple!20]{1}{60}{120}[Parameter\\-isolation \\based]
%     \annulus[purple!30]{1}{120}{180}[Replay\\-based]
%     \annulus[blue!10]{1}{180}{240}[Regularization\\-based]
%     \annulus[blue!20]{1}{240}{300}[Parameter\\-isolation \\based]
%     \annulus[blue!30]{1}{300}{360}[Replay\\-based]
% \end{onion}

\subsubsection{Continual Reinforcement Learning Environments}
\Cref{tab:rlenv-papers} is an extension to the discussion from \Cref{sec:rl_envs} to provide a summary of the frequently used RL environments and their specific characteristics. It is noticeable that most evaluations on Atari had been carried out using rehearsal-based learning algorithms. 

\begin{table}[h!]
    \centering
    \footnotesize
    \rowcolors{2}{gray!25}{white}
    \begin{tabular}
    {|p{3.9cm}|p{1.6cm}|p{0.8cm}|p{4.7cm}|p{1.6cm}|}
        \rowcolor{gray!50}
        \hline
        RL Environment & MDP / POMDP & Text input &  Research work evaluated on this environment & Github resource \\
        \hline
        Atari~\citep{bellemare2013arcade} & MDP \& POMDP & X & ~\cite{rusu2016progressive, kirkpatrick2017overcoming, fernando2017pathnet, schwarz2018progress, ketz2019continual, rolnick2019experience, wang2021afec, atkinson2021pseudo, powers2022cora}  &\href{https://github.com/Farama-Foundation/Arcade-Learning-Environment/tree/master}{Link} \\
        \rowcolor{LightCyan}
        Continual World~\citep{wolczyk2021continual} &MDP&X & ~\cite{wolczyk2022disentangling, ben2022lifelong} & \href{https://github.com/awarelab/continual_world}{Link}\\
        Minigrid~\citep{MinigridMiniworld23} & POMDP & \checkmark & ~\cite{kessler2022same, ben2022lifelong, daniels2022model} & \href{https://github.com/Farama-Foundation/MiniGrid}{Link} \\
        Procgen~\citep{cobbe2019procgen} & MDP& X &~\cite{powers2022cora, powers2022self, ben2022lifelong} & \href{https://github.com/openai/procgen}{Link} \\
        Open AI Gym~\cite{brockman2016openai} & POMDP & X&~\cite{wang2019boosting, kessler2020unclear, kessler2022same, steinparz2022reactive} & \href{https://github.com/openai/gym}{Link} \\
        \rowcolor{LightCyan}
        Meta-World~\citep{yu2020meta}&MDP&X &~\cite{berseth2021comps, caccia2023task}&
        \href{https://github.com/Farama-Foundation/Metaworld}{Link}\\
        CTGraph~\citep{soltoggio2023configurable} & POMDP & X & ~\cite{ben2022lifelong} &\href{https://github.com/soltoggio/CT-graph}{Link} \\
        DeepMind Lab~\citep{beattie2016deepmind} & POMDP&X&~\cite{rolnick2019experience} & \href{https://github.com/google-deepmind/lab}{Link} \\
        Jelly Bean World~\citep{jbw:2020} & POMDP &X &~\cite{steinparz2022reactive, anand2024prediction}& \href{https://github.com/eaplatanios/jelly-bean-world}{Link} \\
        Minihack~\citep{samvelyan2021minihack} &Configurable&\checkmark&~\cite{powers2022cora} & \href{https://github.com/facebookresearch/minihack}{Link} \\
        ALFRED~\citep{shridhar2020alfred} & POMDP&\checkmark & ~\cite{powers2022cora} & \href{https://github.com/askforalfred/alfred}{Link}\\
        % Ai2-thor~\cite{kolve2017ai2} & &~\cite{powers2022cora} &\\
        % Distral~\cite{teh2017distral} & &&& &~\cite{schwarz2018progress} & \\
        \rowcolor{LightCyan}
        Surreal~\citep{corl2018surreal}&Configurable&X& ~\cite{huang2021continual}&
        \href{https://github.com/SurrealAI/surreal}{Link}\\
        % Labyrinth~\cite{jaderberg2016reinforcement}&&X&&3D&\href{}{Link}
        Mazebase~\citep{sukhbaatar2015mazebase}&Configurable&\checkmark&~\cite{sorokin2019continual} &\href{https://github.com/facebookarchive/MazeBase}{Link}\\
        MinAtar~\citep{young2019minatar}&MDP&X&~\cite{anand2024prediction}&\href{https://github.com/kenjyoung/MinAtar}{Link}\\

        % \textcolor{red}{DMControl?}&&&&\\
        \hline
    \end{tabular}
    \caption{Commonly utilized RL environments and corresponding studies. \colorbox{LightCyan}{Colored rows} denote robotic environments}
    \label{tab:rlenv-papers}
\end{table}

\section{Continual Reinforcement Learning in Robotics}
\label{sec:CRL_robo}

The integration of CL in robotics empowers a robot to master multiple tasks instead of relying on task-specific robots~\citep{kalashnikov2021mt}. The aim is to enable the standalone robot to acquire proficiency across a range of tasks, facilitating its adaptation to novel and intricate challenges over time. This mirrors the learning process observed in children. Developmental robotics~\citep{lungarella2003developmental} seeks to create autonomous agents inspired by the developmental trajectory, principles, and mechanisms observed in a child's natural cognitive system. Consequently, these agents should demonstrate the ability to confront varied scenarios while acclimating to unfamiliar environments~\citep{thrun1995lifelong}.

Forward transfer of knowledge and skill is essential in robots, as they are expected to operate efficiently in environments resembling those they were trained on, alleviating the need to train from scratch. One factor contributing to non-stationarity in robotic CL is caused by wear and tear, the gradual deterioration of components due to prolonged use. Other factors include changes in surroundings, modifications to the physical properties of objects handled by the robot or objects in the surroundings, etc. Limitations in computational power, training datasets, time-consuming procedures in data labelling, and many other fundamental and complex issues have delayed the exploration of CL in robotics compared to its application in other domains~\citep{lesort2020continual, ibarz2021train}. Please refer~\cite{lesort2020continual} for a detailed discussion on the state-of-the-art CL approaches in robotics and non-robotics scenarios. Robotics paired with CL can be applied to household robots, autonomous driving systems, and other robots that navigate wide areas to accomplish tasks, where safety remains a critical concern.

The Offline Distillation Pipeline proposed by~\cite{zhou2022forgetting} addresses certain challenges brought forth by off-policy data in RL for CL namely; (i) The limitation in the forward and backward transfer of experience in offline RL problems, even while having access to a populated replay buffer. This issue has been resolved by collecting data across environments and distilling it to a single policy. (ii) The performance drop resulting from training policies with imbalanced datasets containing varying quality and size \wrt to multiple environments is addressed by following the conservative objective in offline RL. Here the learnt policy is made to stay closer to the behavior policy through a KL divergence constraint. The proposed solution is a fusion between an online interaction phase to learn the latest task efficiently using an offline RL algorithm and an offline distillation phase that ensures the current policy stays closer to the behavioral policy. The algorithm has been tested on a bipedal walking robot under varying conditions that introduce non-stationarity. DisCoRL a task agnostic approach introduced by~\cite{traore2019discorl}, is an online method inspired by policy distillation. It uses state representation learning~\citep{lesort2018state} to efficiently compress the inputs (data from a random policy) to a low dimensional embedding using an encoder, subsequently utilized for policy learning. The policy generates data sequences, which are later used for distillation. Addressing challenges in sample efficiency and stability, the approach involved training an omnidirectional robot through domain randomization across various simulated test beds. The trained model demonstrated successful transferability when deployed on a real robot.

A multi-task lifelong learning setting for robotics based on RL has been tested on ten tasks on a Franka Emika Panda robot arm~\citep{xie2022lifelong} where learning new tasks efficiently by reusing past experience was the end goal. A 2x improvement was achieved by learning in a CL manner compared to learning from scratch by involving pre-training on selective past data using a replay buffer and importance weight sampling. An approach intending to teach a robot various motion skills through demonstrations of trajectories without having access to past data was carried out by~\cite{auddy2023continual}. They employed a hypernetwork~\citep{ha2016hypernetworks} to generate weights of a neural ordinary differential equation enabling the recall of longer sequences of skills(26 tasks). For shorter task sequences, a chunked hypernetwork was used. Finally, the learned policy was successfully deployed to a physical robot. 

With the increasing adoption pre-trained visual representations (PVR)~\citep{he2022masked, baevski2022data2vec} for efficient feature extraction, their use has become particularly prominent in supervised and self-supervised learning contexts for downstream tasks. However,~\cite{majumdar2023we} found that existing PVRs perform sub-optimally in embodied AI when evaluated using their benchmark, CortexBench, which consists of 17 tasks focused on diverse robotic manipulation objectives. Their ViT large~\citep{dosovitskiy2020image} model, trained on egocentric videos from seven sources and ImageNet, known as VC-1, has proven to outperform prior PVRs.

Given the increasing prevalence of robotics in everyday activities, the need for an autonomous agent capable of continuous learning and adaptation while maintaining optimal performance in previously mastered tasks is crucial for ensuring efficiency and adaptability in dynamic and evolving environments. Although the existing work~\citep{minelli2023towards, liu2023continual, vodisch2023covio},  in CL for robotics has not reached a perfect level, the area of research is open to explore.
\section{Metrics}
\label{sec:metrics}

Using metrics that accurately reflect the extent of forgetting, the ability to learn new knowledge, and the capacity to leverage past experiences for future learning is crucial in evaluating performance in CL scenarios.

The standard measure of performance for an RL agent is the average expected cumulative reward. Although this metric can help track the agent's progress throughout its learning process, it falls short in evaluating the agent's ability to perform in a CL scenario. This is due to the fact that with diverse tasks, the reward function, complexity \etc, may change. In a CL context, it is crucial to monitor how previous learnings contribute to future ones efficiently (forward transfer), how current learning helps improve previous ones (backward transfer), and how they support preventing catastrophic forgetting.

Below, we discuss in detail the commonly employed metrics in CRL problems and the mathematical derivations. Presenting results as an average across a number of random seeds is a standard practice in RL and CRL experiments due to the instability of the RL algorithms on the network initialization.

\subsection{Evaluation methods for virtual non-robotic and robotic RL environments}
When comparing the performance of a continual learner, it has been the widely accepted act to evaluate it against multi-task (MTL) learner agents having access to all tasks~\citep{caccia2022task}. In such a case, the agent is conscious of the task being presented, and hence its performance is an upper bound for CL agents that are subjected to catastrophic forgetting. 
The manner in which robotic agents are evaluated is different from that of the rest. This is due to the goal-oriented learning nature and the success of achieving pre-defined goals embedded in robot structures and programs~\citep{lesort2020continual}. 
% Using metrics that accurately reflect the extent of forgetting, the ability to learn new knowledge, and the capacity to leverage past experiences for future learning is crucial in evaluating performance in CL scenarios. 
% \begin{table*}[h!]
%     \centering
%      \resizebox{0.90\linewidth}{!}{
%     \Large\addtolength{\tabcolsep}{-0.2pt}
%     \begin{tabular}{r l}
%     $r_{i,j,end}$ & Expected return received for task {i} after being trained on task {j}  \\ 
%     $r_{i,all,max}$ & Maximum expected return gained on task {i} after being trained on all tasks \\
%     \end{tabular}
%     }
%     \label{tab:forgetting_explanation}
% \end{table*}

\Cref{tab:metrics} provides a summary and comparison of performance evaluation mechanisms in place for non-robotic and robotic continual learning setups.

\begin{table}[h!]
\setlength\extrarowheight{1pt}
\centering
\footnotesize
\rowcolors{2}{gray!25}{white}
    % \begin{tabular}{|>{\raggedright\arraybackslash}m{20mm}|m{55mm}|m{55mm}|}
    \begin{tabular}{|p{2cm}|p{5.5cm}|p{5.5cm}|}
    \rowcolor{gray!50}
     \hline
     % \Centering{\textbf{Metric}}&\Centering{\textbf{Non-robotic Environment}}&\Centering{\textbf{Robotic Environment}}\\
    \multicolumn{1}{|>{\Centering\arraybackslash}p{20mm}|}{\textbf{Metric}} 
    & \multicolumn{1}{>{\Centering\arraybackslash}p{55mm}|}{\textbf{Non-robotic Environment}}
    & \multicolumn{1}{>{\Centering\arraybackslash}p{55mm}|}{\textbf{Robotic Environment}}\\
     \hline
        \Centering{Notation}
        &Following~\cite{powers2022cora} \newline$r_{i,j,end}$ - Expected return received for task $i$ after being trained on task $j$, \newline$r_{i,all,max}$- Maximum expected return gained on task $i$ after being trained on all tasks
        &Following~\cite{wolczyk2021continual}, total number of tasks ($N$), current task number (\textit{i}), number of steps trained on (\textit{S}), success rate of task \textit{i} at time \textit{t} (\textit{p\textsubscript{i}}(\textit{t}) $\in$ [0,1]);
        \\
        \hline
        \Centering{Average performance}
            &Average cumulative reward obtained across $T$ steps;
            \begin{equation}
                \textstyle{P_{i,t=T} = \frac{1}{T}\sum^{T}_{t=1}r_{i,i,t}}
            \end{equation}
            &Average success rate across tasks
            \begin{equation}
                \textstyle{P(\textit{t}) =  \frac{1}{N} \sum_{i=1}^{N} \textit{p\textsubscript{i}}(\textit{t})}
            \end{equation}
            \\
        \hline
        \Centering{Forward transfer} 
                  &  Quantifies the aid of past acquired knowledge effectively in future learning situations. This is done by comparing the maximum expected return achieved for the later task i before and after training on task j (i$>$j). 
                  \begin{equation}
                       \textstyle{FT_{i,j} = \dfrac{{r_{i,j,\text{end}}} - {r_{i,j-1,\text{end}}}}{|{r_{i,\text{all},\text{max}}|}}}
                  \end{equation}
                  $FT_{i,j} >0$ - model has gotten better at task i
                & Defined as the normalized area between the training curve (Tr) of the CL agent and the single task expert (STE). \newline FT for task \textit{i} and the average FT are given by; 
                 \begin{equation}
                    \textstyle{FT_{i} = \frac{\text{AUC}_{\text{STE}(i)} - \text{AUC}_{Tr(i)}}{1 - \text{AUC}_{Tr(i)}}}
                \end{equation}
                \begin{equation}
                    \textstyle{FT = \frac{1}{N} \sum_{i=1}^{N} FT_{i}}
                \end{equation}\\
        \hline
        \Centering{Catastrophic forgetting} 
              & This metric is quantified using the expected return of task i that was gained before training on task j and after training on task j (i$<$j). 
             \begin{equation}
                  \textstyle{CF_{i,j} = \dfrac{{r_{i,j-1,\text{end}}} - {r_{i,j,\text{end}}}}{|{r_{i,\text{all},\text{max}}|}}}
                  \label{eqn:catas}
              \end{equation}
              $CF_{i,j} >0$ - model has forgotten task i
              &
              Defined as the drop in performance after training on task \textit{i};
            \begin{equation}
                \textstyle{CF_{i} = \textit{p\textsubscript{i}}(\textit{i}\cdot\textit{S}) - \textit{p\textsubscript{i}}(N\cdot\textit{S})}
            \end{equation}\\
        \hline    
        \Centering{Backward transfer} 
             & Using knowledge gained from the current task to improve a prior learnt task when revisited. This is negative forgetting. From \Cref{eqn:catas}  $\textstyle{CF_{i,j}} < 0$ 
             &Negative forgetting given by $\textstyle{CF_{i}} < 0$\\
            \hline
        \Centering{Continual evaluation} 
             & This metric is displayed on a graph by continuously plotting the evaluation on all tasks periodically while being trained on a new task. Refer~\cite{powers2022cora} for an overview
             & \RaggedRight{A similar evaluation between the continual learner and the single task expert can be observed in~\cite{wolczyk2021continual}.}\\
            \hline
        \Centering{Visual evaluations}
             &\Cref{fig:2env_train_eval} depicts a sequential CL training process across 2 RL environments and the respective evaluation (dashed) curves. As the CRL agent initiates the learning of task 2, a discernible decline in the evaluation for task 1 can be noted. Such visual plots can converse more in comparison to numbers.
             &\RaggedRight{Graphical representation of forward transfer and the transfer matrix are available in~\cite{wolczyk2021continual}}. \\
        \hline
    \end{tabular}
    \caption{Common metrics used to evaluate the performance of CRL agents in non-robotic and robotic environments.}    
    \label{tab:metrics}
\end{table}

\begin{figure}[h!]
     \includegraphics[width=0.9\linewidth,trim={0cm 0cm 0cm 0cm}]{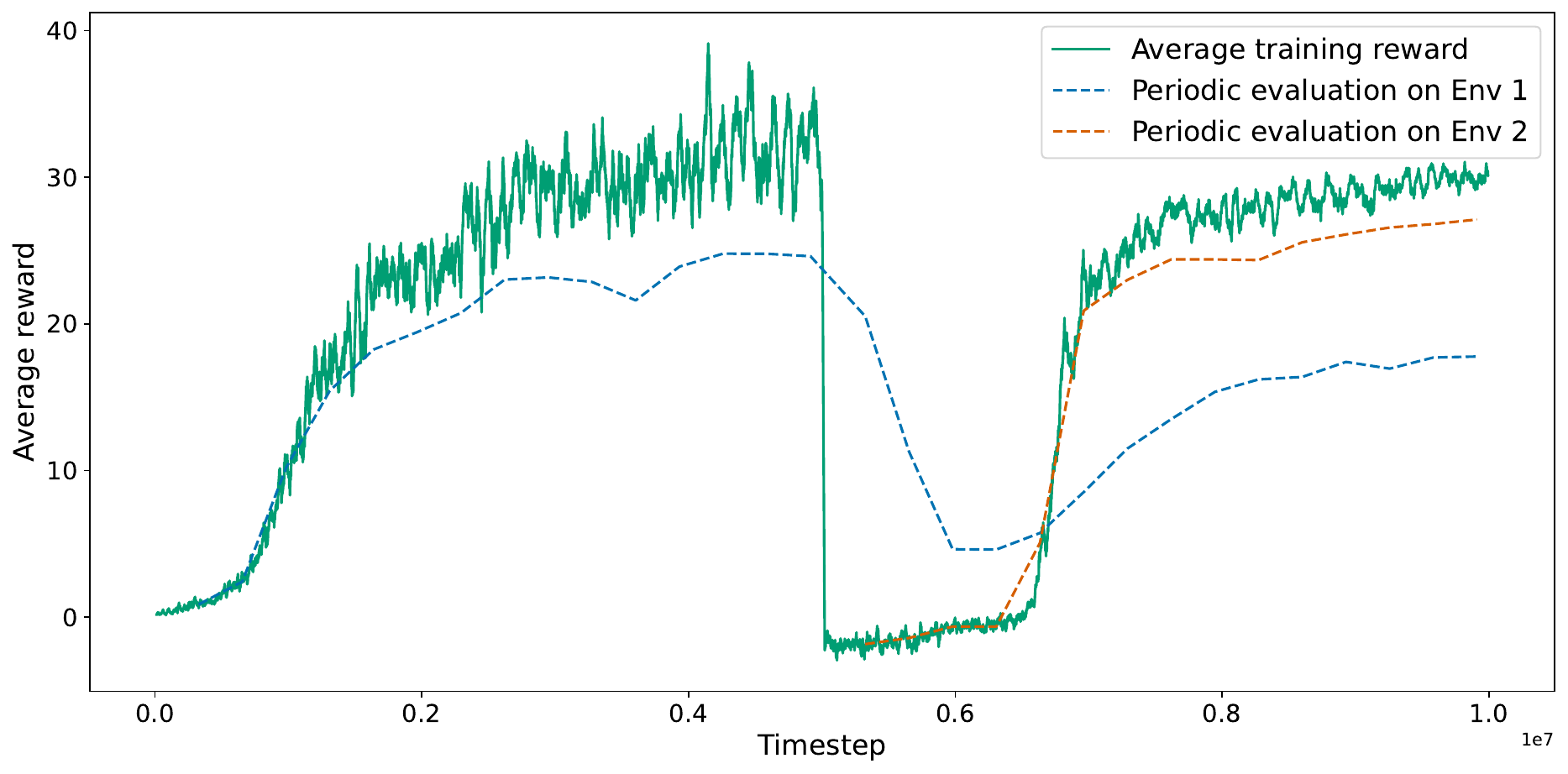}
     \caption{The continual training curve(in green) across 2 environments, each trained for 5M steps. The dashed lines indicate the continual evaluation curves for environment 1 (blue) and environment 2 (orange), respectively.}
     \label{fig:2env_train_eval}
\end{figure}

Apart from the quantitative metrics discussed in \Cref{tab:metrics}, qualitative analysis conducted to showcase diverse measurements of CRL agents include,
\begin{itemize}
    \item Area Under the Curve (AUC) - A frequently used metric representing Forward Transfer~\citep{wolczyk2021continual, ben2022lifelong}
    \item Task success - As defined by robotics environments~\citep{xie2022lifelong, schopf2022hypernetwork}
    \item Reconstruction error - In generative replay-based approaches, the quality of the reconstructions affects the forward transfer of knowledge to the next task~\citep{caselles2018continual}. 
\end{itemize}

\subsection{Proposed Metrics}
In addition to the metrics outlined in \Cref{tab:metrics}, some other metrics that could be considered are skill re-usability in hierarchical CRL setups and efficiency of task-agnostic approaches in identifying new tasks. 

In a hierarchical CRL setup, the agent could be enabled to learn diverse skills as separate tasks. Complex tasks could be solved optimally by meaningfully combining the knowledge obtained through the exposure to these skills. Skill re-usability quantifies the extent to which skills learned in previous tasks can be effectively applied to new tasks without significant degradation in performance. Skill transfer efficiency could be calculated by comparing the model performance ($P$) on new tasks with and without access to previously learned skills.
\begin{equation}
    \text{Transfer efficiency} = \frac{{P}_{\text{with transfer}} - {P}_{\text{without transfer}}}{{P}_{\text{with transfer}}} \times 100\%
\end{equation}

In task-agnostic approaches it is essential for the CL agent to conclude whether the new task at hand is an unseen task or not. The efficiency of this decision-making process is crucial, as it directly influences the responsiveness of the CL system to environmental changes and its ability to swiftly adapt to new tasks. This efficiency can be quantified by assessing the minimum number of steps or data samples required to achieve a certain level of task identification accuracy. Such a metric is vital for ensuring timely adaptation and minimizing delays in the CL process caused by exposure to numerous new tasks.
These metrics are essential for further enriching the evaluation criteria of CRL scenarios. 
% \MH{can you expand upon these here? in particular, explain what each one would measure and why they are important. Try to formulate them mathematically as well}\AZ{expanded the points}

\subsection{Benchmarks}
\subsubsection{Benchmarking Tools}
As important as it is to evaluate the performance of a standalone algorithm, so is comparing it against other similar methods. To bridge the gap in the lack of standard benchmarks for reproducible experiments, some studies have been conducted, facilitating their extension according to user preferences. \Cref{tab:benchmarks} summarizes the currently available benchmarking tools/platforms and their specific characteristics.

\begin{table}[h!]
    \centering
    \footnotesize
    \rowcolors{2}{gray!25}{white}
    \begin{tabular}{|p{2.5cm}|p{2.8cm}|p{2.8cm}|p{3.5cm}|p{1.4cm}|}
    % {|l |c| c|c | c|}
        \rowcolor{gray!50}
        \hline
        \textbf{Benchmarking Tool} & \textbf{Evaluated environments} & \textbf{CL Algorithms} &\textbf{Evaluation metrics} &\textbf{Github resource}\\
        \hline
        CORA~\citep{powers2022cora}& Atari, Procgen, MiniHack, CHORES& EWC, Online EWC, P\&C, CLEAR& Continual evaluation, forgetting, forward transfer& \href{https://github.com/AGI-Labs/continual_rl}{Link}\\
        TELLA~\citep{fendley2022continual}& Atari, Minigrid, Cartpole& Configurable&Performance, catastrophic forgetting, sample efficiency, forward transfer, information retention from previous tasks &\href{https://github.com/lifelong-learning-systems/tella}{Link}\\
        Avalanche-RL~\citep{lucchesi2022avalanche} & Open AI Gym, atari, Habitat& EWC &Episodic return, episodic length, losses etc.&\href{https://github.com/ContinualAI/avalanche-rl}{Link}\\
        Continual World~\citep{wolczyk2021continual}&10 robotics tasks from Meta-World~\cite{yu2020meta}&EWC, MAS, VCL, PackNet, Perfect Memory, A-GEM& Performance, forgetting, forward transfer& \href{https://github.com/awarelab/continual_world}{Link}\\
        COOM~\citep{tomilin2024coom}&Based on ViZDoom~\cite{kempka2016vizdoom}& EWC, MAS, VCL, PackNet, Perfect Memory, A-GEM, ClonEx-SAC& Average performance, forgetting, forward transfer& \href{https://github.com/hyintell/COOM}{Link}\\
        LIBERO~\citep{liu2023libero} (For robotics)&Robotic tasks from Ego4D~\cite{grauman2022ego4d}& Experience Replay, EWC, PackNet& Forward transfer, neghative backward transfer, area under the curve& \href{https://github.com/Lifelong-Robot-Learning/LIBERO}{Link}\\
        \hline
    \end{tabular}
    \caption{A summary of benchmarking tools and their characteristics}
    \label{tab:benchmarks}
\end{table}

Although task sequences in Continual World~\citep{wolczyk2021continual} have not been tested for backward transfer of knowledge, it is a well-suited benchmark capable of addressing this since, in test tasks, the agent is allowed to revisit old tasks learned. 

\href{https://github.com/google-deepmind/bsuite}{Bsuite}~\citep{osband2020bsuite} contains carefully crafted experiments to explore the vital capabilities of RL agents (such as scalability or memory consumption) on a variety of benchmarks and environments. The experiments are tailored to the environments and the number of episodes involved. Experiments have been designed to evaluate agents based on core criteria such as basic learning, stochasticity (robustness to noisy rewards), problem scale (robustness to reward scales), exploration, credit assignment and memory. While this platform allows for studying the baseline performance of both on-policy and off-policy agents, it also assists in understanding the enhancements to standard algorithms via carefully crafted evaluation plots.
This open-source base could be extended to CL scenarios with automated evaluations already provided within the repository. \href{https://github.com/lebrice/Sequoia}{Sequoia}~\citep{normandin2021sequoia} is a recent work focused on providing a unified software program for CL in supervised and RL domains.
The software contains a wide variety of CL methods, RL algorithms, environments, and metrics that could effortlessly be extended and customized according to the needs.

\section{Future Directions}
\label{sec:future}

Amidst the dynamic advancements in CRL, recent efforts have predominantly concentrated on addressing the challenge of catastrophic forgetting. In comparison to the intricate processes occurring within the human brain, there exists more potential for enriching CRL algorithms beyond merely mitigating catastrophic forgetting. This involves facilitating RL agents to systematically assimilate knowledge and subsequently apply it. In this section, we delve into the existing gaps within the literature and explore potential enhancements that could elevate the efficacy of current CRL algorithms.

% \subsection{Challenges}
% \subsection*{Insights from neurocognitive learning}
\textbf{Insights from neurocognitive learning.}
In \href{\ref{lab:replay}}{replay based methods}, partial replay and temporally structured replay methods already functioning in the brain have not been explored in a CL context~\citep{hayes2021replay}. They are known to improve efficiency, allowing better generalization and memory combinations and better stability. 
When employing replay buffers, akin to the memory function in the brain, adopting methods that store fewer samples but effectively capture the essence of encountered tasks aligns well with a CL setup. This is particularly pertinent as the number of saved samples increases linearly with each new task~\citep{robins1995catastrophic}. As discussed by~\cite{sarfraz2023study} in the brain, dendritic segments aid in discovering the effect of information encoding in context signals. This opens research in CL to decide which information could be helpful in sequential learning tasks. Backpropagation has also been criticized in CL models since the corresponding synapses in the brain function in a unidirectional manner, making feedforward and feedback connections distinct. Research resembling localized learning taking place in the brain has also not been explored. 

% Reservoir sampling~\cite{vitter1985random} has been attested to balanced inclusion of recent samples to overcome bias between old vs new samples thus improving CL~\cite{sarfraz2023study}.
% \newline

\textbf{Limitations in learning environments.}
Many of the available learning platforms were built to assess the performance of single expert agents, making them unsuitable for CL problems unless they are adapted and repurposed to suit the scenario~\citep{lomonaco2021avalanche, normandin2021sequoia}. The lack of practical and representative benchmarks has slowed the development of vital algorithms in CRL. One solution is to use video games as learning environments by changing certain aspects of them to enable non-stationarity in the data distribution~\citep{machado2018revisiting, cobbe2019procgen}. Although many recent works have identified these limitations and have designed appropriate benchmarks on which CRL algorithms could be tested on, there is still room to improve them with respect to limitations in state complexity, task continuity, first-person-view offerings, lightweight on computational resources, providing realistic renderings (image-based) and provide standard metrics and relevant evaluations.

When reflecting on how humans interact, text input also offers a form of knowledge in addition to visual input. This is another avenue of research to be tested in the CRL context. Some environments~\citep{MinigridMiniworld23} have already provided the means to such resources. 

\textbf{Efficient feature learning methods from RL environments.}
RL agents make informative decisions based on the state representation that the policy receives. These state representations can be of varying dimensionality, resulting from complex or simpler inputs. When highly rich inputs such as image frames (\eg, Atari games) are received, the feature encoder network is responsible to provide the RL agent with meaningfully encoded representations. In a CL setup, such encoders need to also maintain parameters responsible to hold knowledge across all tasks. Currently, there has been some research produced on this, such as~\cite{caselles2018continual, caselles2021s}, but they have only been tested on very primitive RL environments. An empirical study by~\cite{ostapenko2022continual} depicts that transfer, forgetting, task similarity and learning are dependant on the characteristics of input data more than the CL algorithm.
Some of the works on robotics such as~\cite{raffin2019decoupling, ren2021experimental} have looked into incorporating vision to robotic tasks and to efficiently perform these tasks using state representation learning. These have merely been explored in a CL setup.  An emerging area of research is the application of PVRs for feature extraction in a CRL context. Given that most existing PVRs are trained on natural images, adapting them to the unique dynamics of RL environments, such as those in video games, presents significant challenges. This adaptation necessitates a thorough reassessment of their performance and suitability in these contexts.

% \newline
\textbf{Evaluation metrics.}
Due to the dynamic nature of learning evaluating CRL algorithms using the traditional metrics may not fully grasp the model's performance across evolving tasks. In addition to the current metrics, rate of adaptability, generalization ability, forgetting rate, skill transferability and the ability in skill composition can add greater value to evaluate performance of CRL agents in non-robotic and robotic based environments.

\textbf{Learning without task boundaries.}
Some of the more general problems to overcome are task-agnostic learning methods. Many solutions out there require explicit specification on the task switch during training and testing phases. In real-world scenarios, the transition from one task to another is rarely noticed. Agents in such environments require mechanisms to autonomously identify when the current task/ data distribution changes and decide to consolidate knowledge without human intervention. This also mirrors the learning process within the brain. 

% \newline
% \textbf{}
% Under constraint driven methods, the hardest problem is finding a way to update model parameters in a way that least affects prior learnings. Although many methods have been introduced in the supervised setting, not all of them function as expected when applied on RL algorithms. With respect to parameter-isolation approaches, there are more opportunities to test this setup on CRL scenarios.  In the context of lifelong learning, handling data variability during inference, especially when encountered data significantly differs from the training data, remains an unexplored area~\cite{kudithipudi2022biological}. This is because the data used to train the model is noise free in order to retrieve optimal performance.

\textbf{Network initialization issues.}
Another problem that arises with RL-related algorithms is related to network initialization. This also causes the experiments to be difficult to reproduce, matching the performance from the original work. The adapted solution has been to report performance across five or more seeds.~\cite{colas2018many} has discussed this problem in depth with a statistical perspective. Masking methods have shown a positive transfer of useful knowledge~\citep{ben2022lifelong}, but careful and selective initialization of the base network holds the key to success in this setup.

Handling data variability during inference, especially when encountered data significantly differs from the training data, remains an unexplored area~\citep{kudithipudi2022biological}. This is because the data used to train the model has been cleaned to be noise-free in order to retrieve optimal performance.

\textbf{Evaluating backward transfer.}
One evident drawback in the existing research trajectory is the insufficient emphasis on enhancing backward knowledge transfer. While catastrophic forgetting has been extensively studied, there has been limited exploration of methods to enhance the retention and transfer of knowledge in the reverse direction~\citep{lin2022beyond}.

\textbf{Order of task sequence.}
It is also obvious that, most solutions available to overcome challenges in CRL, have handpicked and specifically designed task sequences making it highly unrealistic to practical scenarios.~\cite{wolczyk2021continual} discusses in an extended study how the performance and forward transfer of knowledge significantly differs when the task order is changed. CL methods tested included EWC~\citep{kirkpatrick2017overcoming} and PackNet~\citep{mallya2018packnet}. This emphasizes how skill acquisition in an order could be beneficial or detrimental.  

% transformer/prompt based solutions?
% \textbf{Effect of pre-trained models in CL}
% As pre-trained models on large-scale datasets become increasingly available, it has paved way to investigate how CL could benefit from such models. These models can effectively serve as feature extractors and hence support in the process of knowledge transfer. The empirical study on this topic by~\cite{ostapenko2022continual} depicts that transfer, forgettting, task similarity and learning are dependant on the characteristics of input data more than the CL algorithm. The same study can be carried out in CRL to learn the underlying factors that could uplift the performance of these algorithms \AZ{complete}

\textbf{Issues in robotic CL.}
In the context of CL for robotics, a much more complex set of problems needs to be addressed due to the continuous interaction with a physical environment. Since this interaction is expensive with respect to data collection and sample inefficiency, most solutions have opted for simulation-based setups or training with static datasets in an offline manner. Certain works assume the possibility of storing all samples collected~\citep{xie2022lifelong}, although this could pose scalability challenges. The potential to use generative algorithms in this case would be a good solution to reduce the storage overhead. In sim-to-real transfer, this deviation between the simulation training and the actual environments results in tangible performance drops~\citep{zhao2020sim}.

To conclude, when examining how humans acquire new concepts, it becomes evident that there is a gradual progression of knowledge, starting from fundamental facts (simplest concepts) and extending to the most sophisticated ones (complex concepts). It is infeasible for a biologist to suddenly become an expert in astronomy unless the learning process follows a systematic and incremental approach. In the same manner in the context of CRL, it is important to focus more on developing CL algorithms that can handle complex environments by initially familiarising with basic ones. The process should align with data and resource efficient learning methods encountering longer task sequences. Backward transfer of knowledge and stable learning mechanisms are of high necessity to improve the state of the art work. Compared to CL approaches in supervised learning, RL requires more influential and generally applicable algorithms to learn continually.

\section{Conclusion}
The field of CL has thus evolved to incorporate different forms of learning. With the heavy influence of mammalian learning systems, much of the current work has been successful in enabling RL agents to learn amidst non-stationary data distributions like humans and animals do. Although lately, much focus has been given to mitigating catastrophic forgetting, concepts from the cognitive brain still offer many other problems and respective approaches to be developed and tested in lifelong learning settings. Addressing the open problems discussed above could elevate the existing methodologies to solve more complex problems in RL contexts with high certainty. The end goal of CRL is to develop adaptable agents capable of offering solutions generalized and applicable to real-world scenarios. We hope this survey helps navigate the research area that aims to produce AI agents with general intelligence.

\vskip 0.2in
\bibliography{references}
\bibliographystyle{apalike}

\end{document}